\documentclass[10pt,twocolumn,letterpaper]{article}

\usepackage[pagenumbers]{cvpr}

\usepackage{graphicx}
\usepackage{amsmath}
\usepackage{amssymb}
\usepackage{booktabs}
\usepackage{multirow}  
\usepackage{array}     

\usepackage[pagebackref,breaklinks,colorlinks]{hyperref}

\usepackage[capitalize]{cleveref}
\usepackage[utf8]{inputenc} 
\DeclareUnicodeCharacter{FF0C}{,}

\crefname{section}{Sec.}{Secs.}
\Crefname{section}{Section}{Sections}
\Crefname{table}{Table}{Tables}
\crefname{table}{Tab.}{Tabs.}

\def\cvprPaperID{*****} 
\def\confName{CVPR}
\def\confYear{2025}

\begin{document}

\title{Maximizing the Impact of Deep Learning on Subseasonal-to-Seasonal Climate Forecasting: The Essential Role of Optimization}

\author{Yizheng Guo$^{}$\thanks{Equal contribution}\\
Tsinghua Shenzhen International Graduate School\\
{\tt\small orogeny@qq.com}
\and
Tian Zhou$^{}$\footnotemark[1]\\
Alibaba DAMO Academy\\
{\tt\small tian.zt@alibaba-inc.com}
\and
Wangyi Jiang\\
University of the Chinese Academy of Sciences\\
{\tt\small jiangwanyi@mail.iap.ac.cn}
\and
Bo Wu\\
University of the Chinese Academy of Sciences\\
{\tt\small  wubo@mail.iap.ac.cn}
\and
Liang Sun\\
Alibaba DAMO Academy\\
{\tt\small liang.sun@alibaba-inc.com}
\and
Rong Jin\\
Alibaba DAMO Academy\\
{\tt\small rongjinemail@gmail.com}
}
\maketitle

\begin{abstract}
Weather and climate forecasting is vital for sectors such as agriculture and disaster management. Although numerical weather prediction (NWP) systems have advanced, forecasting at the subseasonal-to-seasonal (S2S) scale, spanning 2 to 6 weeks, remains challenging due to the chaotic and sparse atmospheric signals at this interval. Even state-of-the-art deep learning models struggle to outperform simple climatology models in this domain. This paper identifies that optimization, instead of network structure, could be the root cause of this performance gap, and then we develop a novel multi-stage optimization strategy to close the gap. Extensive empirical studies demonstrate that our multi-stage optimization approach significantly improves key skill metrics, PCC and TCC, while utilizing the same backbone structure, surpassing the state-of-the-art NWP systems (ECMWF-S2S) by over \textbf{19-91\%}. Our research contests the recent study that direct forecasting outperforms rolling forecasting for S2S tasks. Through theoretical analysis, we propose that the underperformance of rolling forecasting may arise from the accumulation of Jacobian matrix products during training. Our multi-stage framework can be viewed as a form of teacher forcing to address this issue. Code is available at \url{https://anonymous.4open.science/r/Baguan-S2S-23E7/}
\end{abstract}

\section{Introduction}
Weather and climate forecasting encompasses various time scales, including nowcasting, short and medium-range weather forecasting, subseasonal-to-seasonal forecasting, and climate forecasting~\cite{subseasonal2seasonalprediction,lorenz1963deterministic}. These predictions are crucial for social management, agriculture, disaster preparedness, and other sectors, reflecting our understanding of Earth's status as humanity's sole home~\cite{osti_1657487,gmd-16-3407-2023,doi:10.1073/pnas.2216158120,ThePhysicsofHeatWavesWhatCausesExtremelyHighSummertimeTemperatures}. Among these, subseasonal-to-seasonal (S2S) forecasts can predict major natural disasters~\cite{white2017potential,vitart2012subseasonal} such as droughts and floods~\cite{doi:10.1073/pnas.2216158120}, which play a significant role in disaster preparedness and policy-making. It is well recognized that accurate forecasts for the S2S regime—specifically, 2 to 6 weeks ahead—remains an operational challenge due to the chaotic nature of the atmospheric system
and the differing temporal scale mechanisms involved compared to short-term weather forecasting~\cite{Tellus, Estimating}.

\begin{figure}[t]
  \vskip -0.2in
  \centering
  \begin{subfigure}[b]{0.49\linewidth} 
    \centering
    
    \includegraphics[width=1\linewidth]{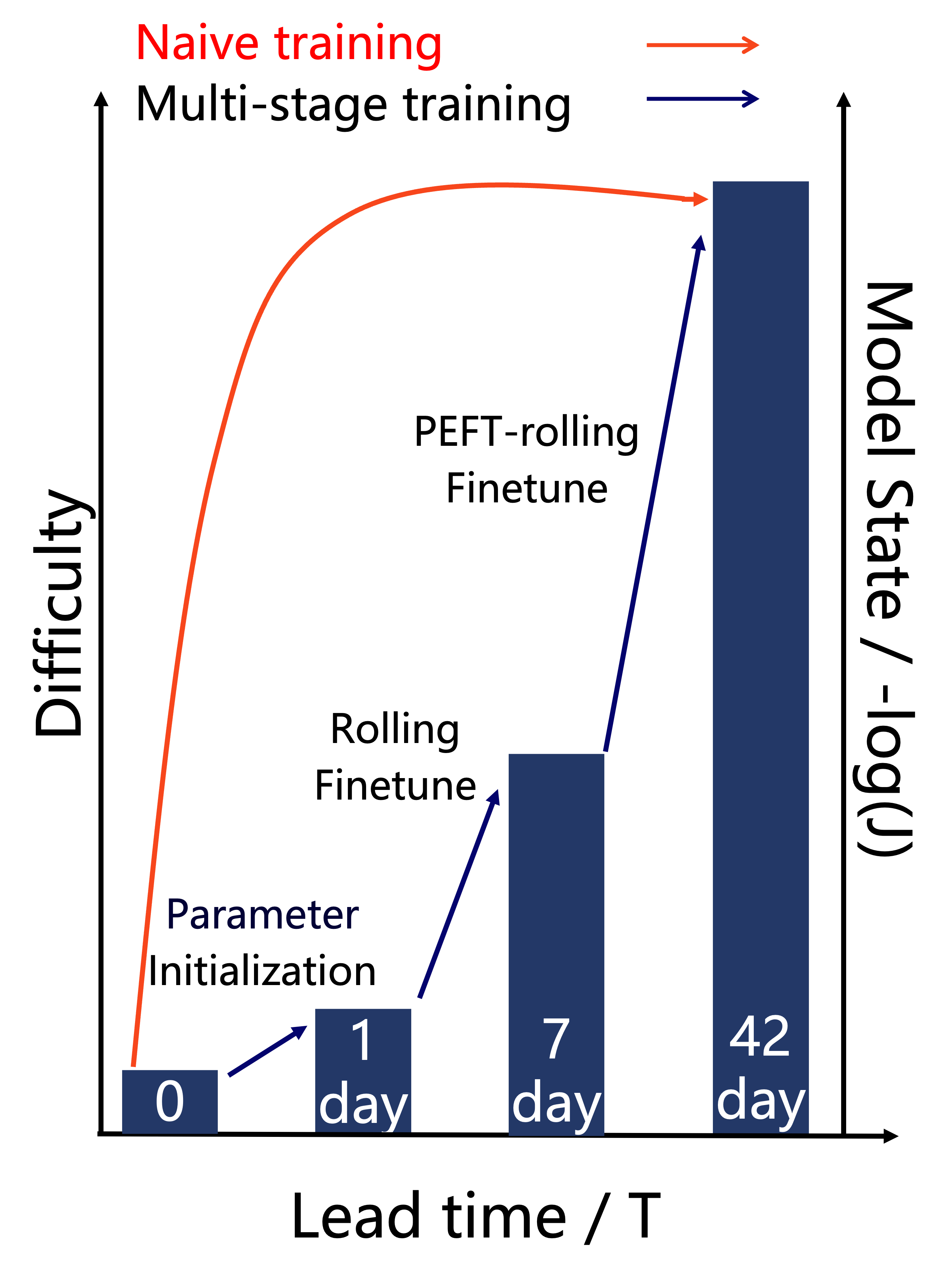}
  \end{subfigure}
  \hfill
  \begin{subfigure}[b]{0.50\linewidth} 
       \includegraphics[width=1\linewidth]{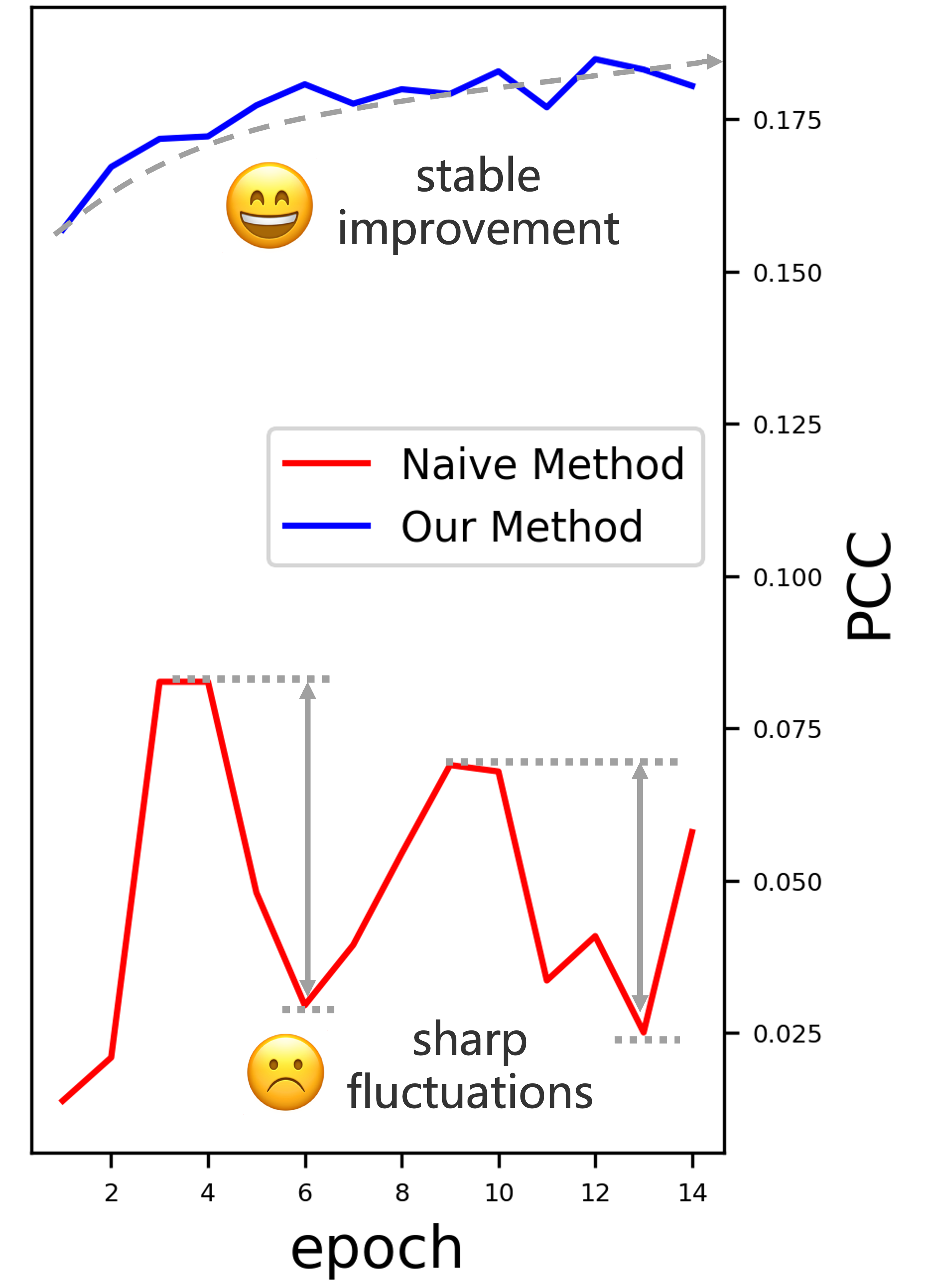}
   \end{subfigure}
   \caption{  \textbf{Left}: comparison of the training processes of the multi-stage method with the naive method. The naive method, while increasing $T$, exhibits a significant discrepancy in the model state, leads to gradient divergence. Our method gradually increases $T$ while simultaneously reducing the discrepancy in model states at each stage, thereby lowering the training difficulty. \textbf{Right}: training stability comparison between naive training method and our multi-stage training method.}
   \label{fig:Naive and Our}
   \vskip -0.2in
\end{figure}

Currently, commonly used weather and climate forecasts rely on Numerical Weather Prediction (NWP) systems~\cite{bauer2015quiet,lorenc1986analysis,coiffier2011fundamentals}, which are based on solving physical ordinary differential equations (ODEs) for thermodynamics, fluid flows, etc. In meteorology, weather signals decay and eventually disappear over time, typically ranging from 1 hour to 14 days, while climate signals gradually emerge over a period of 2 to 4 months. In between, the S2S forecasting occurs on a time scale typically ranging from 14 days to 42 days (2 weeks to 6 weeks). The extreme sparsity of signals in this interval makes it very challenging for physical-based NWP systems to produce reliable forecasts. In addition, both the huge computational cost and limited modeling capacity for the physical world significantly constrain the application of NWP methods in practice. In recent years, machine learning methods—particularly deep learning models~\cite{Nguyen2023ClimaXAF,pathak2022fourcastnet,Chen2023FuXiAC} for weather and climate forecasting—have emerged, significantly reducing computational costs and improving the ability of directly learning from data. In weather prediction tasks, these models~\cite{bi2022pangu,doi:10.1126/science.adi2336,Chen2023FengWuPT,Bodnar2024AuroraAF}, have already outperformed state-of-the-art complex numerical weather prediction systems, mostly on nowcasting and short/medium-range weather forecasting. 

But the story becomes very different when coming to S2S forecasting: recent studies~\cite{nathaniel2024chaosbench,mouatadid2023adaptive} have shown that as the forecasting range approaches the S2S time scale, almost all state-of-the-art (SOTA) deep learning weather forecasting models, including ClimaX~\cite{Nguyen2023ClimaXAF}, Pangu-weather~\cite{bi2022pangu}, GraphCast~\cite{doi:10.1126/science.adi2336}, and FourCastNet v2~\cite{Bonev2023SphericalFN, pathak2022fourcastnet}, are indistinguishable from unskilled simple climatology model, which averages weather statistics from many years as the prediction. 
The key observations from these studies lead to two important questions as follows. 
\begin{itemize}
\item \textbf{Why do these models, with millions of learnable parameters, fail to outperform simple, non-parametric climatology baselines in S2S tasks?}
\end{itemize}
Due to the nature of sparse signals at S2S time scale, we believe these deep models for S2S forecasting usually suffer from overfitting, especially when complex neural networks are applied blindly. The climatology method, detailed in the metrics section and using the average of past true signals, serves as a robust baseline to address this overfitting. Rather than focusing on designing more complex neural networks, we approach the challenge of S2S forecasting through the lens of optimization. Furthermore, our optimization strategy is crafted to tackle the inherent sensitivity of S2S weather forecasting to both initial conditions (IC) and boundary conditions (BC), as noted in previous studies~\cite{Tellus,Estimating,lorenz1963deterministic}. These factors are often regarded as two extremes that intricately affect the S2S signal.
By adopting a multi-stage teacher forcing framework, our experiments show a significant improvement in both Pearson Correlation Coefficient (PCC) and Temporal Correlation Coefficient (TCC) skills, while maintaining the same backbone model structure. Moreover, we improved the performance of the leading subseasonal to seasonal (S2S) forecasts by the European Centre for Medium-Range Weather Forecasts (ECMWF-S2S), achieving enhancements of over \textbf{19-91\%}. This demonstrates that deep learning methods, with appropriate design of optimization strategy, can be a very effective approach for S2S forecasting.

Our second question is about the comparison between direct forecasting and rolling forecasting, i.e., 
\begin{itemize}
\item \textbf{Is direct forecasting truly more accurate than rolling forecasting for S2S weather predictions?}
\end{itemize}
According to \cite{nathaniel2024chaosbench}, S2S forecasting methods can be classified into two categories, direct forecasting and rolling forecasting. Direct forecasting employs time-stamp-like prompts and directly predicts weather outputs at a specified horizon, as exemplified by ClimaX~\cite{Nguyen2023ClimaXAF}. In contrast, for rolling forecasting, models only predict weather outputs for the next time unit, and model outputs are iteratively fed back into the system to extend the prediction time frame, such as GraphCast~\cite{doi:10.1126/science.adi2336}, FuXi-S2S~\cite{chen2023fuxis2s}, and the hybrid approach used in Pangu-weather~\cite{bi2022pangu}. Studies in \cite{nathaniel2024chaosbench} suggest that, in S2S forecasting tasks, the accuracy of rolling forecasts consistently falls short compared to direct forecasts, regardless of the base models. 
Our findings present a counterargument. The relatively limited performance of rolling forecasting methods compared to direct forecasting approaches is caused by the model optimization problem. Our study shows that a multi-stage teacher forcing framework improves the accuracy of the rolling forecasting methods—measured by Pearson and Temporal Correlation Coefficients—by over \textbf{30\%} compared to direct forecasting. This improvement is independent of the backbone model, rendering it significantly more effective for S2S tasks than direct forecasting.

To motivate the multi-stage teaching force framework, we first provide in \cref{fig:Naive and Our} the empirical observations of training a 42-day rolling forecasting model. We can observe a large fluctuation in training errors -- a phenomenon of training instability that is largely due to the accumulation of prediction errors at each step. In the proposed optimization strategy, we introduce two mechanisms to improve the training stability, i.e., 
teacher forcing for the intermediate steps to mitigate the accumulated prediction errors, and multi-stage learning to reduce the difficulty of end-to-end training by learning easy tasks and difficult ones later. The combination of both essentially makes our training significantly more stable for S2S tasks compared to one-stage optimization methods for rolling forecasting, as illustrated in \cref{fig:Naive and Our}.

\section{Related Work}
S2S prediction primarily relied on numerical weather prediction~(NWP) methods from meteorological centers such as the UK Meteorological Office~(UKMO)~\cite{33}, the National Centers for Environmental Prediction~(NCEP)~\cite{34}, the China Meteorological Administration~(CMA)~\cite{35}, and the European Centre for Medium-Range
Weather Forecasts~(ECMWF)~\cite{36}. Among them, ECMWF's weather forecasts are widely recognized as the most accurate, and they publicly provide the forecast results to society. However, NWP models are extremely costly. With the development of deep learning, models such as Pangu-weather~\cite{bi2022pangu}, GraphCast~\cite{doi:10.1126/science.adi2336}, FourCastNet v2~\cite{Bonev2023SphericalFN}, FuXi~\cite{Chen2023FuXiAC} and ClimaX~\cite{Nguyen2023ClimaXAF} have started to emerge in the field of meteorology. These methods replace the solutions of ODE equations with a fully end-to-end learning approach, achieving better model parallelism and reducing inference costs. Also, methods like ClimODE~\cite{verma2024climode} use deep networks to construct ODE-type architectures, which similarly improve parallelism. However, in the S2S domain, the performance of the above models has been suboptimal. FuXi-S2S~\cite{chen2023fuxis2s} has recognized this issue and employs Variational Autoencoder (VAE~\cite{kingma2013auto})-like methods to enhance the model's accuracy in long-term predictions. However, it addresses the issue with a specialized architectural design, lacking a general perspective. In particular, its generative framework is difficult to replicate and sensitive to training nuances, making it challenging for future work to build upon.

The optimization of the rolling prediction model can be approximated as an optimization of either ultra-deep models or ultra-long RNN models. In \cite{bartlett2018gradient}, an analysis was conducted on the convergence issues of deep linear models and they derived the necessary conditions for convergence. The teacher forcing~\cite{hess2023generalized} method optimizes the training of RNN by controlling the proportion of gradient propagation. Building on these works, we continued to derive the issues related to the rolling prediction model and proposed a multi-stage training approach to reduce the optimization difficulty.

\section{Methodology}
\label{sec:method}

\begin{figure*}
  \centering
  \begin{subfigure}{0.48\linewidth}
    \centering
    \includegraphics[width=1\linewidth]{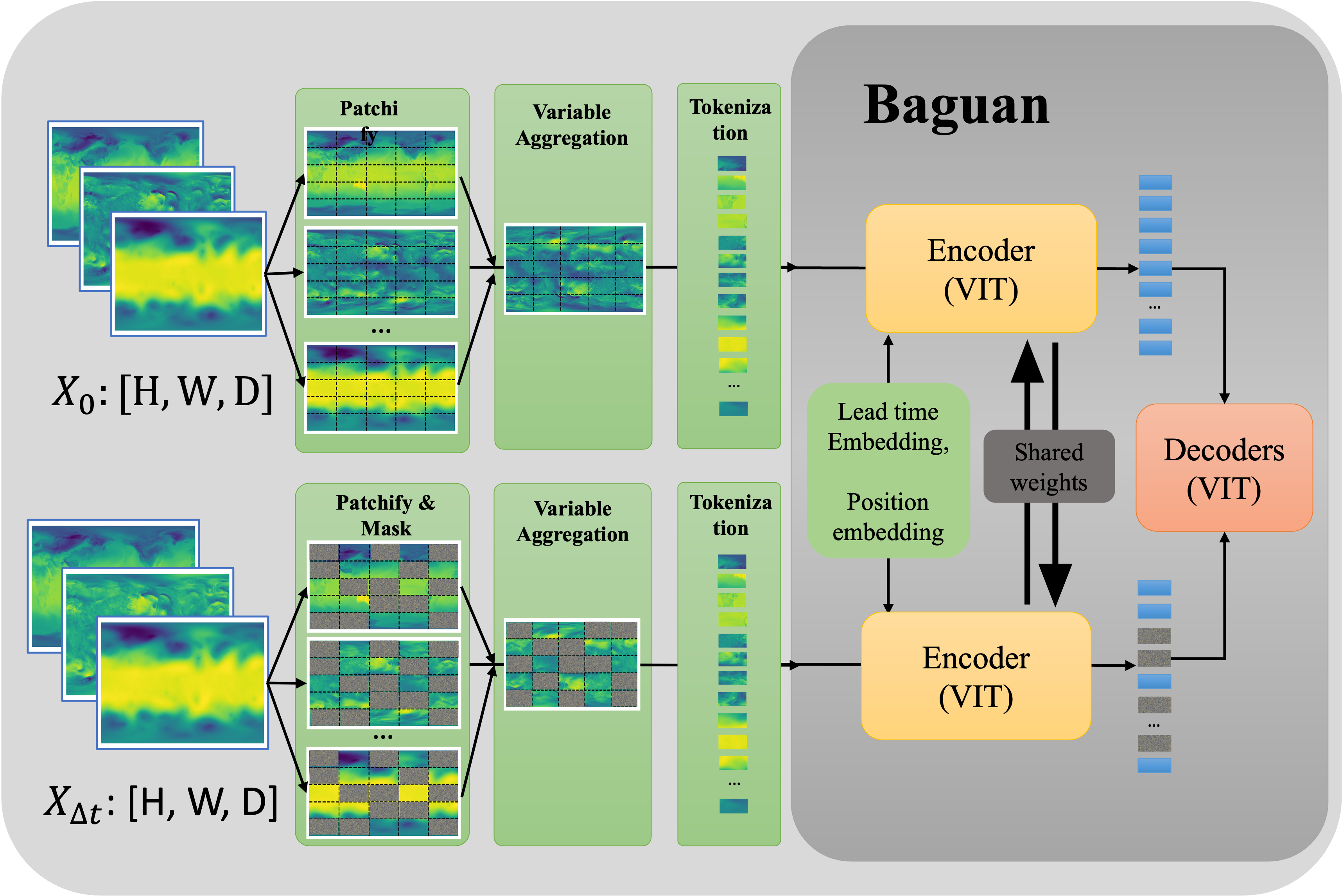} 
    \label{fig:Baguan}
  \end{subfigure}
  \hfill
  \begin{subfigure}{0.51\linewidth} 
    \centering
    \vspace{10pt}
    \includegraphics[width=1\linewidth]{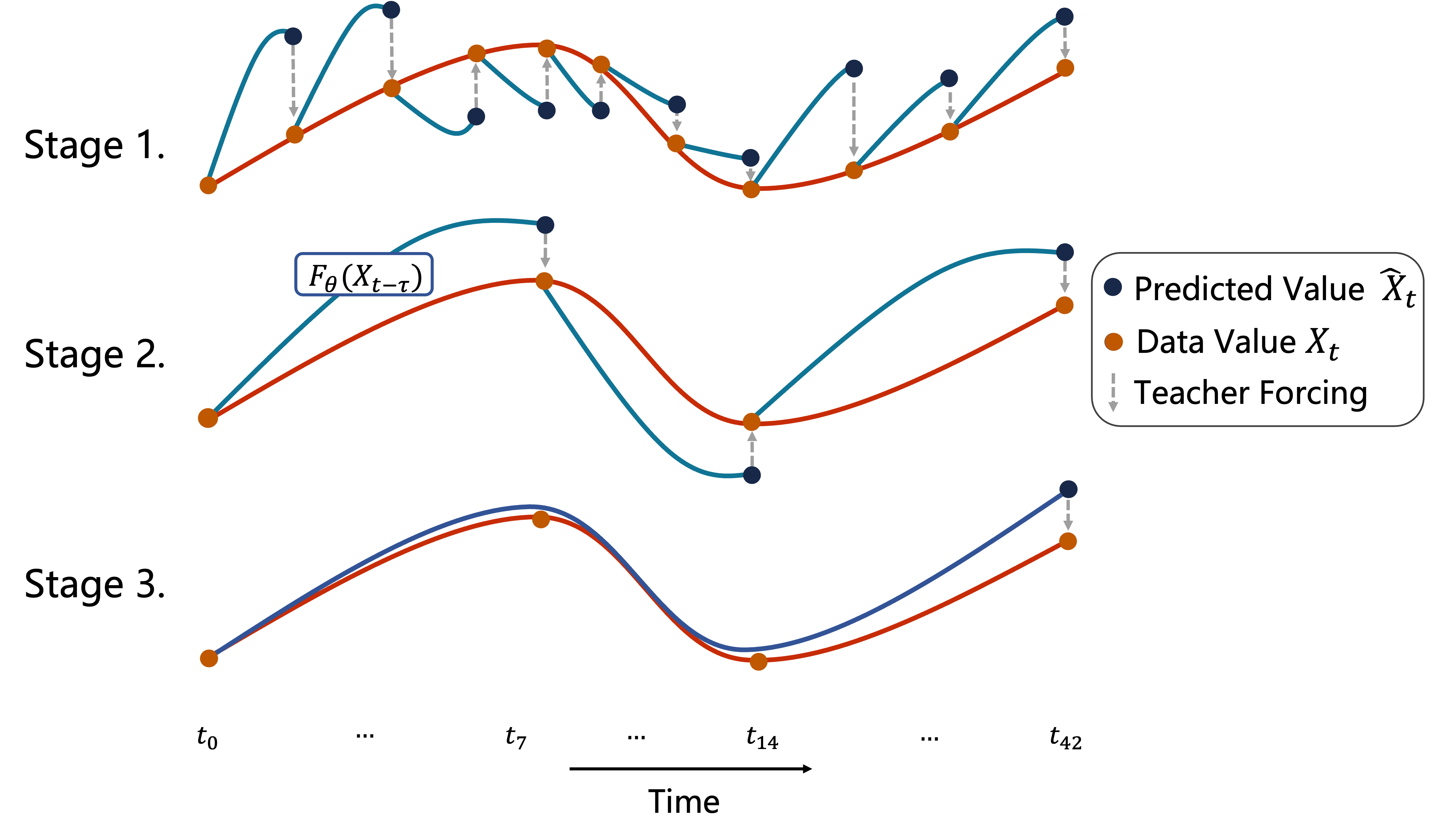}
    \label{fig:Teacher forcing}
  \end{subfigure}
  \caption{Principle of multi-stage progressive learning. \textbf{Left}: structure of our base model Baguan\cite{buguanwebsite}. Baguan employs a Siamese MAE method\cite{gupta2023siamese} to pre-train a ViT-structured weather forecasting model. \textbf{Right}: Demonstration of the Teacher Forcing\cite{hess2023generalized}. Teacher forcing substitutes the observed data values $X_t$ for the predicted values $\hat{X}_t$, thereby interrupting the multiplicative path of $J$. We gradually extend $T$ by controlling the frequency of teacher forcing, therefore enabling long-term rolling optimization of the model.}
  \vskip -0.2in
  \label{fig:main-strategy}
\end{figure*}

\begin{figure}[th]
  \centering
   \includegraphics[width=0.8\linewidth]{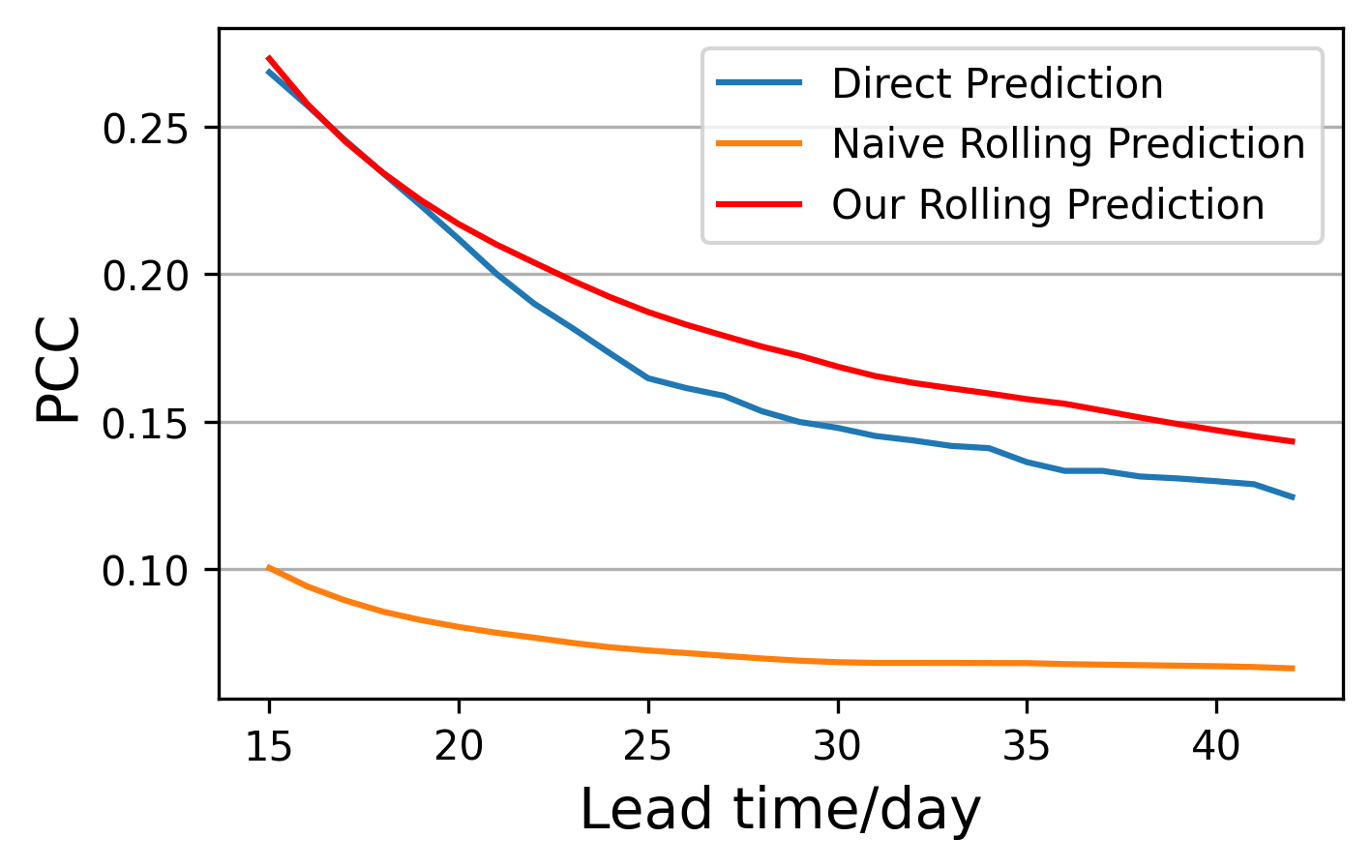}
   \caption{ Comparison between direct prediction and different rolling prediction. In direct prediction, a separate model was employed for each lead time. The naive rolling method loses the model's predictive capability during long-term rolling, whereas our optimized approach effectively addresses this issue.}
   \label{fig:direct and rolling}
   \vskip -0.2in
\end{figure}

As the starting point, our empirical studies suggest that rolling forecasting is more effective than direct forecasting for S2S tasks, as illustrated in Figure~\ref{fig:direct and rolling}, which is opposite to the conclusion from the recent study~\cite{nathaniel2024chaosbench}. We believe it is due to the suboptimal optimization of previous weather forecasting models for long-term rolling prediction. Below we will first provide a simple analysis that motivates the development of a novel optimization strategy, and then present our multi-stage teacher forcing optimization framework in detail.

\subsection{Motivation for Better Optimization Strategy for S2S Forecasting: a Simple Analysis}
Despite its advantage over direct forecasting, rolling forecasting suffers its own problem: the prediction errors made on each step can be accumulated over the time, leading to significant instability in training illustrated in \cref{fig:Naive and Our}. To better understand the accumulation effect of the rolling prediction model, we formalize the rolling prediction model as an RNN (Recurrent Neural Network) given as follows\cite{hess2023generalized}:
\begin{equation}
\hat{X}_t = F_\theta(\hat{X}_{t-1}, s_{t}),
\label{eq.1}
\end{equation}
where $\hat{X}_t$ is the model prediction (or model state) at time $t$, $s_{t}$ is additional external inputs at time $t$, and $\theta$ is the model parameter. Let $\ell(\hat{X}_t, X_t)$ be the loss function comparing ground truth observations $X_t$ and model prediction $\hat{X}_t$. Then the overall loss function is $L(\theta) = \sum_{t=1}^T \ell(\hat{X}_t, X_t)$. 
Define the Jacobian of $F_{\theta}(\hat{X}_{t-1}, s_t)$ w.r.t. $\hat{X}_{t-1}$: 
\begin{equation}
    J_t := \frac{\partial \hat{X}_t}{\partial \hat{X}_{t-1}} = \frac{\partial F_\theta (\hat{X}_{t-1}, s_t)}{\partial \hat{X}_{t-1}}. 
\end{equation}
Following the idea of Backpropagation Through Time (BPTT)\cite{rumelhart1986learning, werbos1990backpropagation}, we can write the gradient $\nabla_{\theta}L(\theta)$ as 

\begin{eqnarray*}
\lefteqn{\nabla_{\theta}L(\theta) = \sum_{t=1}^T \nabla_{\theta} \ell(\hat{X}_{t}, X_t)} \\
& = & \sum_{t=1}^T  \nabla_{\theta}F_{\theta}(\hat{X}_{t-1}, s_t) \frac{\partial \ell(\hat{X}_t, X_t)}{\partial \hat{X}_t}\\
& = & \sum_{t=1}^T \left(\frac{\partial F_{\theta}(\hat{X}_{t-1}, s_t)}{\partial \theta} +  \nabla_{\theta}\hat{X}_{t-1} J_t \right)\frac{\partial \ell(\hat{X}_t, X_t)}{\partial \hat{X}_t} \\
& = & \sum_{t=1}^T \left(\sum_{j=1}^t \partial_{\theta}F_{\theta}(\hat{X}_{j-1}, s_j)\prod_{k=j+1}^t J_k\right)\frac{\partial \ell(\hat{X}_t, X_t)}{\partial \hat{X}_t} \\
& = & \sum_{t=1}^T \partial_{\theta}F_{\theta}(\hat{X}_{t-1}, s_t)\left(\sum_{j=t}^T \left[\prod_{k=j+1}^T J_k\right] \frac{\partial \ell(\hat{X}_j, X_j)}{\partial \hat{X}_j}\right).
\label{eq: loss gradiant}
\end{eqnarray*}
As indicated by the above expression, any local adjustment of parameter $\theta$, i.e., $\partial_{\theta} F_{\theta}(\hat{X}_{t-1}, s_t)$, can be amplified through the product of Jacobian matrices $\prod_{k=j+1}^T J_k$, clear evidence of accumulation effect. Hence, the training stability, i.e., the stability of RNN orbit $\hat{X}_1, \ldots, \hat{X}_T$, is decided by the maximum Lyapunov exponent of Jacobian matrix product along the orbit: 
\begin{equation}
    \lambda_{\text{max}} := \lim_{T \to \infty} \frac{1}{T} \log \left\| \prod_{r=0}^{T-2} J_{T-r} \right\|_2,
\end{equation}
where $\left\| \cdot \right\|_2$ denotes the spectral norm.  Previous research~\cite{mikhaeil2022difficulty} has demonstrated that the condition $\lambda_{max} > 0 $ will inevitably lead to diverging loss gradients when training RNNs on chaotic time series using gradient descent based algorithms. To control $\lambda_{max}$, there are two obvious approaches: moderate sequence lengths $T$ or smaller $J$ for all time. For the rolling prediction model, a rolling length of 42 days is required for the final results regardless; however, shorter sequence lengths can be used in the intermediate steps of training. The matrix $J_t$ is tied to the initial state of the model, and it is very large for randomly initialized parameters. However, through convergent stepwise training, we can optimize the model for a smaller $J$. Combining both ideas, we can formulate a feasible optimization pathway, which involves gradually reducing $J_t$ through incremental training with increasing values of $T$.

Motivated by the above analysis, we propose a multi-stage training method that gradually increases the rolling length (sequence length) in each training stage to constrain $J_t$ under the final length. In particular, we employ a teacher forcing approach, where observed data values are substituted for predicted values at varying frequencies during training to control the rolling length $T$. This approach helps prevent the model from diverging loss gradients in the optimization space. The overall idea of the proposed multi-stage teacher forcing method for training is illustrated in \cref{fig:main-strategy}. We note that the proposed optimization strategy is closely related to curriculum learning~\cite{bengio2009curriculum}, as by gradually increasing the rolling length, we effectively increase the difficulty of training. Finally, the training stability issue of rolling forecasting is also related to the optimization of deep linear network~\cite{bartlett2018gradient}, where the deeper the network is (or more rolling times), the more restricted condition we have for the initial solution in order to obtain a reliable convergence. More details about this line of analysis can be found in \cref{appendix:proof}.

\subsection{A Multi-Stage Teacher Forcing Framework for Optimization}

First, we decompose the task of S2S forecasting into multiple steps. To develop a long-term rolling daily average forecasting model, it is necessary to fit the model as a daily average forecasting model, then a rolling prediction model, and lastly a long-term rolling model. This corresponds to the progression of $T$ from $1$ to $7$ and then to $42$.

In the first stage, we fine-tune the hourly forecast pre-trained model to daily forecast model. The base model we employed is Baguan's 1.4-degree version~\cite{buguanwebsite}, a Siamese MAE method pre-trained ViT-based model for weather forecasting. The architecture comprises an encoder with 8 transformer layers, each having a hidden dimension of 1024, and a decoder with 4 layers maintaining the same hidden dimension. It adheres to the original structure of the Siamese MAE as described by \cite{gupta2023siamese}. Additionally, we use a patch size of 4x4 for processing the input. In this stage, we aim to obtain a highly accurate daily forecasting model, which serves as the base model for subsequent fine-tuning. At this step, 
$T$ is minimal, making the convergence easier. To rapidly reduce $J$, we employ a large learning rate and perform full fine-tuning.

In the second stage, we use a 7-day lead time (corresponds to $T=7$) to train the model's rolling prediction capability. After 7 rounds of rolling predictions, all predicted results are used to compute the loss against the observed values. At this stage, the rolling length is comparatively reasonable with an optimized starting point, so we allow all parameters to be trainable with a small learning rate. 

In the third stage, we fine-tune the model for the 42-day rolling prediction task. In this stage, the rolling depth reaches its maximum value, but we have already obtained better initial parameters through two phases of training, thereby controlling $J$. Through experiments, we have found that training with full fine-tuning leads to instability and offers limited improvements in results. To manage the training challenges arising from the increased rolling length, we employed parameter-efficient fine-tuning (PEFT) to control the number of trainable parameters, thereby restricting the optimization direction of the model. We used a standard adapter~\cite{houlsby2019parameter} approach, employing 6 sets of adapters during the rolling process, with one set utilized every 7 days. The implementation details are described in \cref{appendix:adapter}. Each adapter set comprises 4\% of the total model parameters, and during training, all parameters except for the adapters are frozen. Through three stages of progressive training, the model can be consistently optimized to achieve a better rolling prediction model.
\begin{figure}[t]
  \centering
   \includegraphics[width=0.8\linewidth]{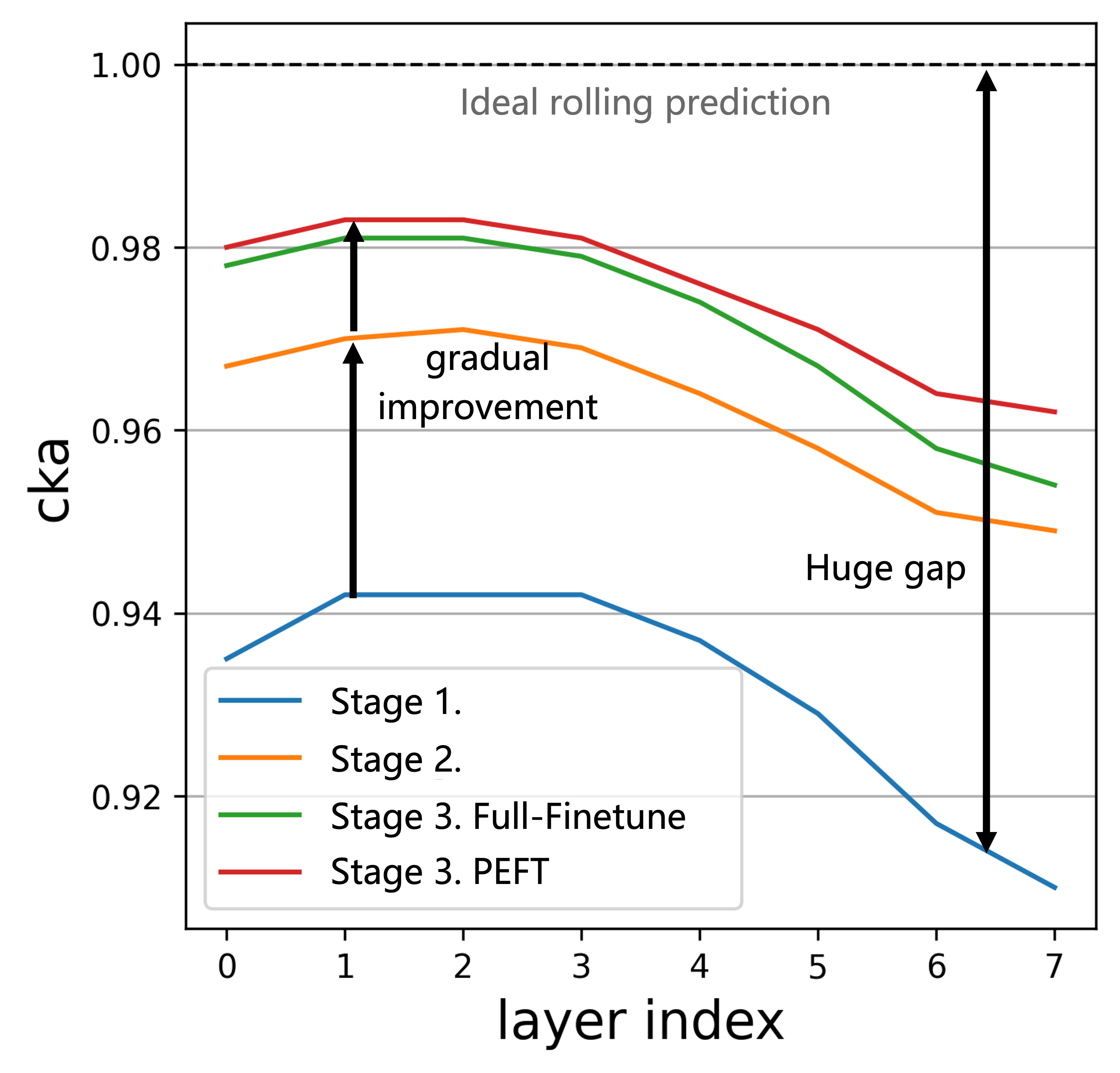}
   \caption{ Performance of multi-stage training compared with ideal rolling prediction. We use CKA to calculate the similarity between the model features $F_{\theta,feat}(X_{41})$ and model features obtained after rolling predicting 41 times with data from day 0 $F_{\theta,feat}(F_{\theta}^{(41)}(X_{0}))$. In an ideal rolling prediction, the two features should be perfectly aligned, which means that CKA equals 1. A higher CKA indicates stronger information continuity, allowing the model to achieve better performance in long-term rolling predictions.}
   \label{fig:onecol}
   \vskip -0.2in
\end{figure}

We employed Centered Kernel Alignment (CKA)~\cite{huh2024platonic, kornblith2019similarity}, a recently proposed method for assessing feature similarity, to evaluate the model's rolling prediction performance at various training stages. As demonstrated in Figure~\ref{fig:onecol}, we compared the CKA curves of the rolling predictions with those from last-day direct predictions across different model stages. The results showed that, with progressive training, the CKA values gradually improved. This suggests that the model's rolling prediction capability increasingly aligns with the ideal scenario, where the input is adjusted based on actual observations rather than predicted values. Comparing the full fine-tuning in the third stage with PEFT, we found that PEFT achieved a higher CKA score and exhibited less decline in CKA at the model's endpoint.

In S2S tasks, meteorologically, the focus is not on the direct outcomes but rather on the anomalies relative to climatology. Therefore, the commonly used MSE loss may not be suitable for this task. Instead, we utilized two anomaly-based loss functions, including latitude-weighted MSE loss and latitude-weighted PCC loss. The simple average of these two loss functions is treated as the final loss. Formally, we use use $A$ to denote anomaly values, and $C$ to denote the observed climatology, and denote $A_n=X_n - C$. 
Then the \text{Atitude-weighted anomalies MSE loss} is defined as
\begin{equation}
  L_{\text{AMSE}} = \frac{1}{N} \sum_{i=1}^{N} w(\phi_i)(A_i - \hat{A}_i)^2.
  \label{eq:atitude:loss}
\end{equation}
\noindent
And the \text{Latitude weighting PCC loss} is defined as
\begin{equation}
  L_{\text{PCC}} = 1 - \frac{\sum_{i=1}^{N} w(\phi_i) (A_i \hat{A}_i) }
  {\sqrt{\sum_{i=1}^{N} w(\phi_i) A_i^2 \times \sum_{i=1}^{N} w(\phi_i) \hat{A}_i^2 }},
  \label{eq:latitude:loss}
\end{equation}
where $\phi_i$ denotes the latitude at position $i$, $w$. 
\section{Experiment}
\begin{figure*}[th]
    \centering
    \includegraphics[width=0.75\linewidth]{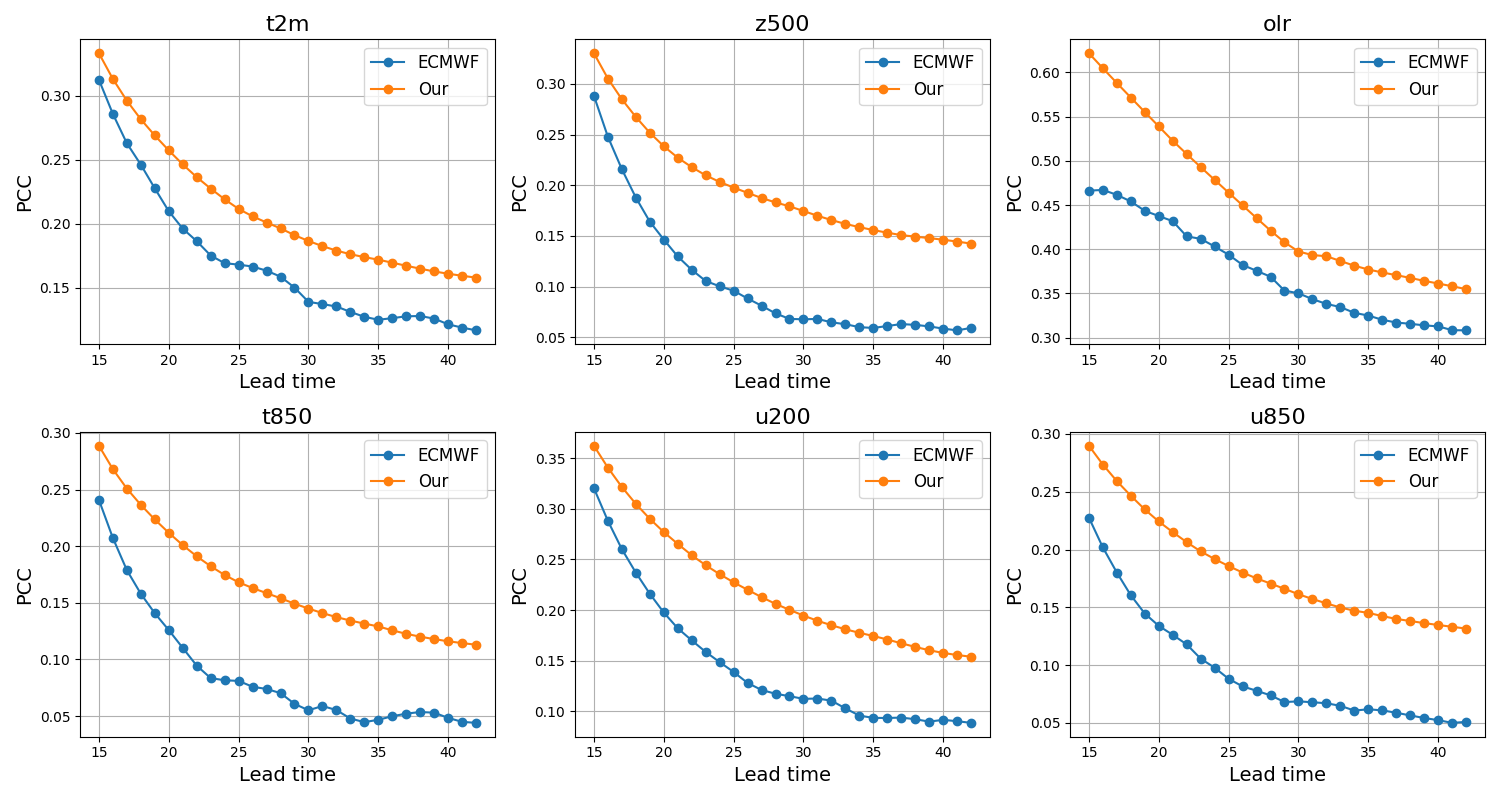} 
    \caption{PCC comparsion between ECMWF-S2S and Our model. Our model surpassed the state-of-the-art NWP systems (ECMWF-S2S) by over \textbf{19-91\%}.}
    \label{fig:main-pcc}
    \vskip -0.2in
\end{figure*}

\subsection{Settings}
\subsubsection{Model Settings}
In the experiments, we used ERA5~\cite{hersbach2020era5} as the observations, which offers a comprehensive record of the global atmosphere by integrating physical models and observational data for correction. The training data includes daily data from 1979 to 2014 and the testing data consists of data from 2015 to 2018. In the main experiments, we used a data resolution of 1.40625 degrees for both latitude and longitude, resulting in a global feature map size of [128, 256]. For rolling prediction, we used the observed value $X_0$ to roll forward and predict $\hat{X}_1$ to $\hat{X}_{42}$. We employed the AdamW optimizer with $\beta_{1}=0.9, \beta_{2}=0.99$.  In the three training stages, the peak learning rates were set at $1e-4, 2e-6, 2e-6$ while the minimum learning rates were $1e-4, 1e-6, 1e-6$, respectively. The model structure we used is Baguan and we also utilized its pre-trained parameters for weather forecasting. The training for each stage converged in approximately 20 epochs and was conducted using NVIDIA A800 GPUs, consuming approximately 32, 192, and 384 GPU hours, respectively. The rolling step increases the per-step training time linearly.

The ECMWF S2S reforecasts (hindcasts) are generated dynamically using the most recent model version available at the time of forecast production. In our study, we utilize the ECMWF S2S reforecasts produced from model cycle C47r3. These reforecasts include initialization dates spanning over a period of 20 years, from January 3, 2002, to December 29, 2021. We also use the period from 2015 to 2018 as the testing set and the climatological mean is computed for the period from 2002 to 2021.

\subsubsection{Evaluation Metrics}
In the experiments, we evaluate the PCC (Pearson Correlation Coefficient) and TCC (Temporal Correlation Coefficient) metrics.
The calculation of PCC is
\begin{equation}
  PCC(\tau) = \frac{1}{B} \sum^{B} \frac{\sum_{i}^{N}w(\phi_i)(A_{i} * \hat{A}_{i}^{\tau})}{\sqrt{(\sum_{i}^{N}w(\phi_i)A_{i}^{2}) * (\sum_{i}^{N}w(\phi_i)(\hat{A}_{i}^{\tau})^{2})}}.
  \label{eq:PCC}
\end{equation}
Here, $B$ denotes the size of the test set, and $\tau$ represents the predicted lead time.
The calculation of TCC is 
\begin{equation}
  TCC(\tau) = \frac{1}{N} \sum^{N}_{i} w(\phi_i) \frac{\sum_{b}^{B}(A_{i,b} * \hat{A}_{i,b}^{\tau})}{\sqrt{(\sum_{b}^{B}A_{i,b}^{2}) * (\sum_{b}^{B}(\hat{A}_{i,b}^{\tau})^{2})}}.
  \label{eq:TCC}
\end{equation}

The two metrics evaluate the similarity of results from both temporal and spatial perspectives. Notably, the calculation of outliers is influenced by the climatology $\textit{C}$. In our experiments, we used a highly accurate climatology approach by calculating the average values for each day of the year across all years, and we applied a rolling average centered on the current day, encompassing the preceding and following 5 days in total. For the climatology of the model predictions and ERA5 data, we utilized the calculations based on ERA5 data and ECMWF employs its own climatology. Using a longer average or fixed time period can weaken the climatology signal, making the results appear inflated.

In the experiments, our model inputs and performs rolling predictions for ten variables: 2m temperature (t2m), geopotential at 500hPa (z500), u component of wind at 200hPa (u200), u component of wind at 850hPa (u850), outgoing longwave radiation (olr), temperature at 850hPa (t850), v component of wind at 200hPa (v200), v component of wind at 850hPa (v850), specific humidity 850hPa (sp850) and total precipitation (tp). The forecast starting points include all days throughout the year. Considering the importance of the signals, we primarily report the first six parameters.

\subsection{Comparison with ECMWF-S2S}
\begin{figure*}[th]
    \centering
    \includegraphics[width=0.75\linewidth]{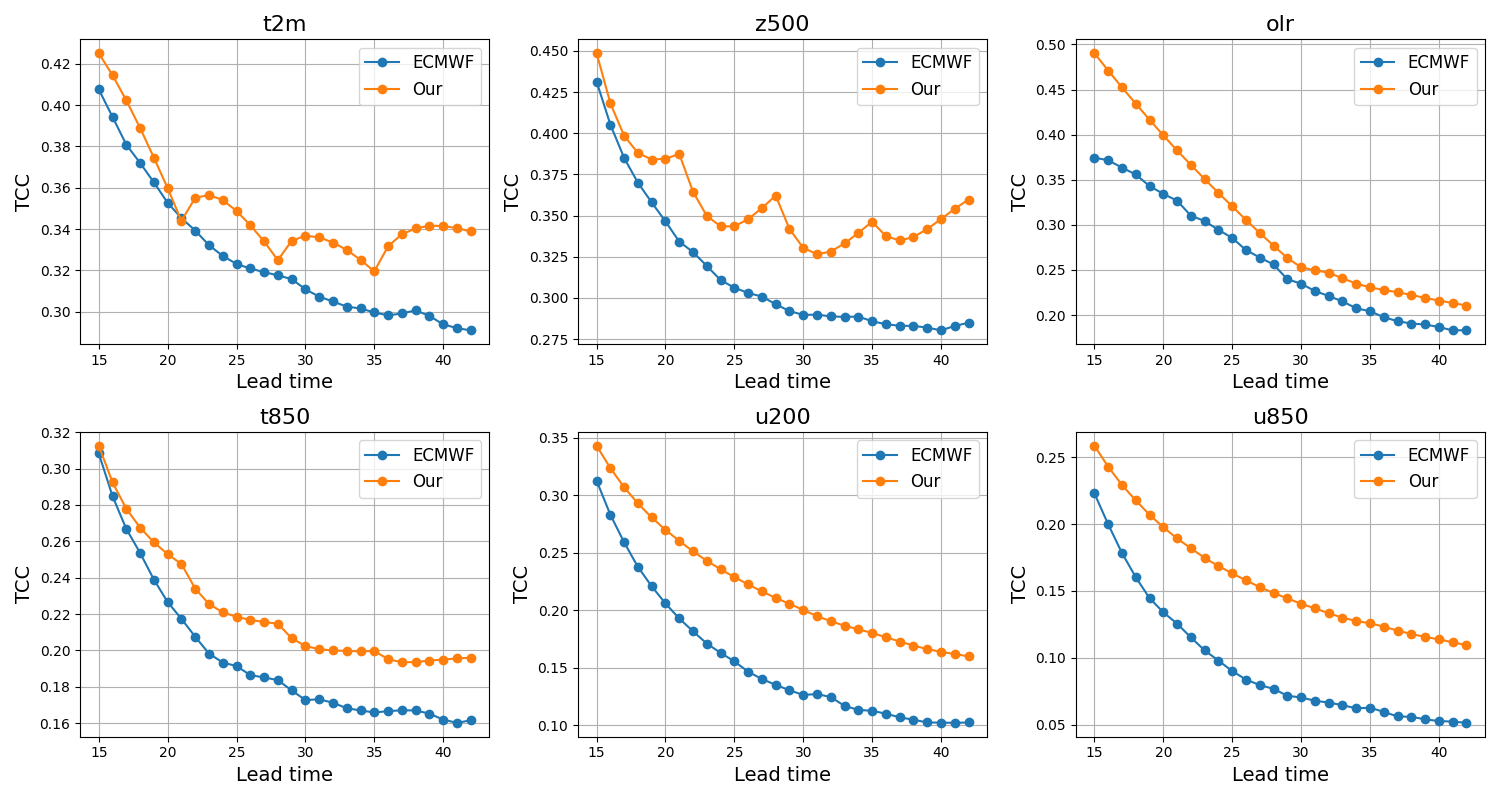} 
    \caption{TCC comparsion between ECMWF-S2S and Our model.}
    \label{fig:main-tcc}
    \vskip -0.2in
\end{figure*}
ECMWF-S2S is currently recognized as the state-of-the-art forecasting model at the global S2S scale. We primarily compare the PCC and TCC results over the time period of 15 to 42 days, corresponding to the 3rd week to the 6th week. In the \cref{fig:main-pcc}, we report the results for t2m, z500, t850, olr, u200, and u850, which are important at the S2S scale. It can be seen that our results show a significant improvement compared to the ECMWF at the S2S scale. Over the entire 2 to 6-week period, improvements remain steady, with an average increase exceeding 35\%. Notably significant enhancements are observed in z500, t850, u200, and u850. At the same time, the results from ECMWF show instability during long prediction times. Among them, our method exhibits a certain 7-day periodicity in the TCC for t2m and z500, which may be attributed to the six sets of adapters, with each set accounting for a 7-day period, in the third stage. Additionally, the visualization samples are available in \cref{appendix:Visualization}, and our model trained with 5.625-degree data outperforms ECMWF, as demonstrated in \cref{appendix:5.625}.

\subsection{MJO Forecast}
\begin{figure}[th]
     \vskip -0.1in
    \centering
    \includegraphics[width=0.7\linewidth]{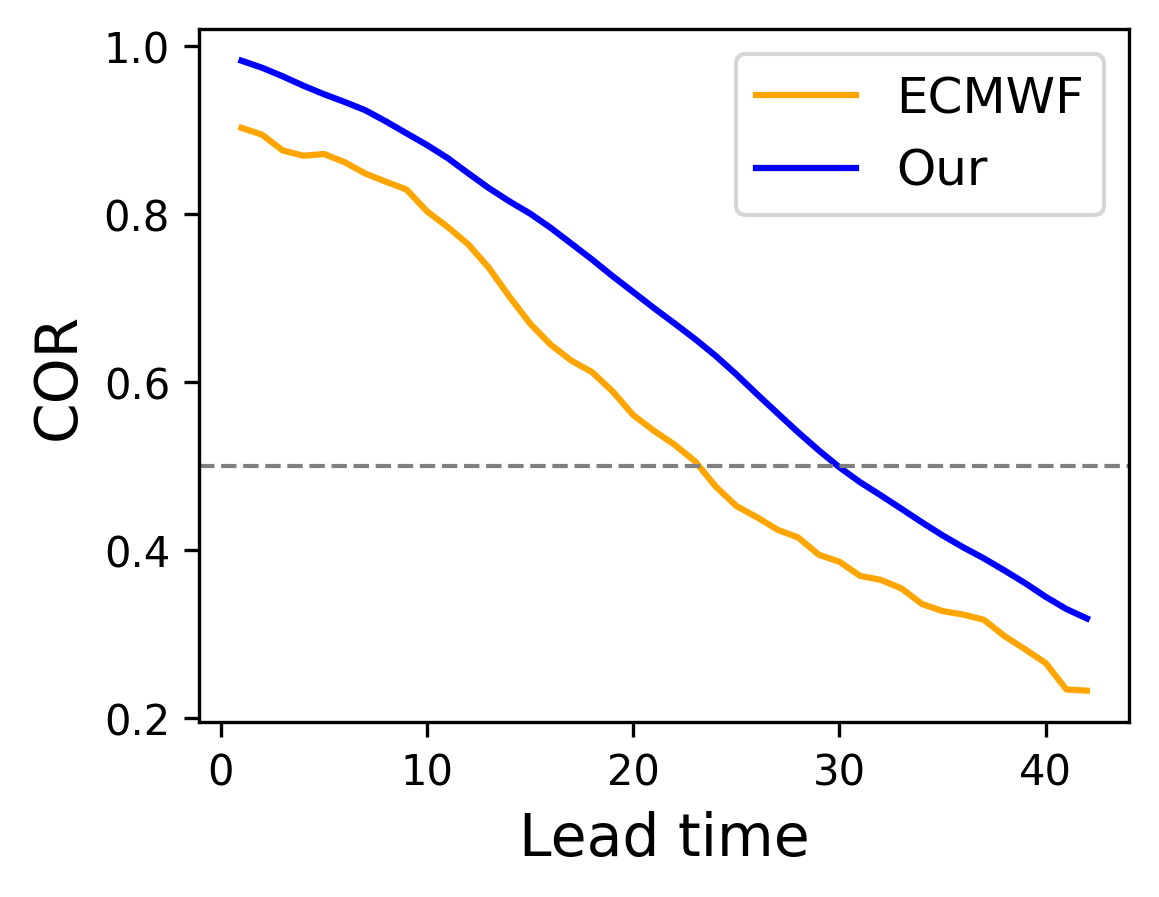} 
    \caption{Comparison of globally-averaged and latitude-weighted RMM bivariate Correlation (COR).}
    \label{fig:mjo}
    \vskip -0.2in
\end{figure}

The calculation of the real-time multivariate MJO (RMM) index \cite{wheeler2004all} is based on the tropical signals of u200, u850, and OLR. To evaluate the MJO forecasting skill, we utilized the bivariate correlation coefficient (COR). The average MJO COR results for 2015 are presented in \cref{fig:mjo}. In the field, a COR threshold of 0.5 is typically used to indicate a skillful MJO forecast. Our model significantly enhances the accurate forecast lead time from \textbf{23 days to 30 days}, exceeding the performance of the ECMWF-S2S model.

\section{Ablation Study}
We conducted extensive ablation studies to validate the effectiveness of our approach. To optimize the budget, all experiments in this section are conducted using the 5.625-degree model. The reported PCC is the average PCC of 6 variables: t2m, z500, olr, t850, u200, and u850.

\subsection{Comparison with Naive Method}
\begin{figure}[th]
 \vskip -0.1in
    \centering
    \includegraphics[width=0.68\linewidth]{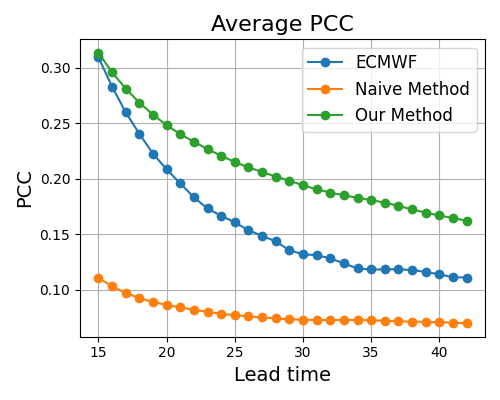} 
    \caption{PCC comparison between naive method and our method.}
    \label{fig:naive pcc}
    \vskip -0.2in
\end{figure}

The naive method directly fine-tunes the model for a 42-day rolling prediction without adding intermediate stages. This method is the simplest approach for training a rolling prediction model. However, the 42 rolling iterations result in excessive depth in training, and the initial state is insufficient to guide the model towards convergence. As shown in \cref{fig:naive pcc}, our method shows a significant improvement in average PCC compared to the naive method.

\subsection{Multi-stage Progressive Learning for Different Backbones}
To validate our proposed Multi-stage progressive learning architecture, we conducted experiments on other backbones. We tested two common models, UNet\cite{ronneberger2015u} and ViT\cite{dosovitskiy2020image} with last stage full fine-tuning. The results are summarized in \cref{tab:backbone}. Detailed results can be found in \cref{appendix:backbone}. The deeper model ViT results in a higher training difficulty compared to the UNet, which leads to worse results with the Naive method. Meanwhile, our method can better leverage the capabilities of a deeper model and result in better performance. Due to budget constraints, we did not test all state-of-the-art weather forecasting models. However, our versatile framework is expected to significantly enhance these models.
\begin{table}
  \centering
   \vskip -0.1in
  \scalebox{0.8}{
  \begin{tabular}{ccc|c}
    \toprule
    Avg. PCC & Our Method & Naive Method & Depth \\
    \midrule
    ViT & 0.1168 & 0.0807 & 51*42 \\
    UNet & 0.1116 & 0.0915 & 15*42 \\
    \bottomrule
  \end{tabular}
  }
  \caption{PCC comparison between our method and naive method with different backbones.}
  \label{tab:backbone}
  \vskip -0.2in
\end{table}

\subsection{Progressive Improvement of Multi-stage Training}
\begin{figure}[th]
    \centering
    \includegraphics[width=0.68\linewidth]{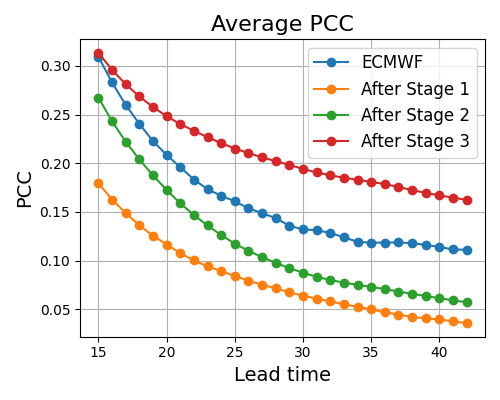} 
    \caption{PCC comparison between multi-stages.}
    \label{fig:3stage ablation}
    \vskip -0.2in
\end{figure}

We validated the effectiveness of the multi-stage training through experiments. By comparing the results of the models trained in the first, second, and third stages, as shown in \cref{fig:3stage ablation}, we can observe that the multi-stage training progressively improves the metrics. Furthermore, after the second stage of training, the model gradually approaches ECMWF, and by the end of the third stage, it surpasses ECMWF.

\subsection{Comparison with Last Stage Full Fine-tune}
\begin{figure}[th]
    \centering
    \includegraphics[width=0.68\linewidth]{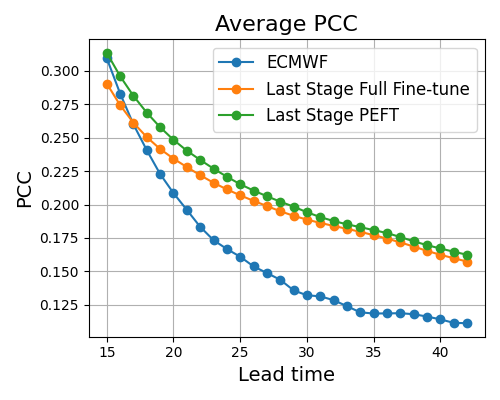} 
    \caption{PCC comparison between full fine-tuning and PEFT in the last stage.}
    \label{fig:peft pcc}
    \vskip -0.2in
\end{figure}

In the final stage, the model's rolling predictions extend from 7 to 42 iterations, eliminating the use of intermediate teacher forcing for training adjustments. At this stage, the model exhibits some rolling prediction capabilities; however, its performance over longer time spans remains suboptimal. To address this, we implemented a Parameter-Efficient Fine-Tuning (PEFT) method. As illustrated in \cref{fig:peft pcc}, utilizing the PEFT method in the final stage leads to enhanced results. The primary reason for this improvement is that by freezing most of the parameters, parameter oscillation is minimized, enabling the model to converge more effectively.

\subsection{Scaling Law in S2S}
\begin{figure}[th]
    \centering
    \includegraphics[width=0.68\linewidth]{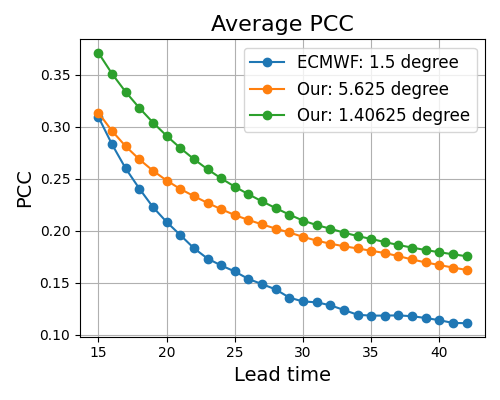} 
    \caption{PCC comparison between 5.625-degree model and 1.40625-degree model.}
    \label{fig:degree}
    \vskip -0.2in
\end{figure}

We observe the scaling law in S2S tasks. We conducted comparative experiments between the 5.625-degree model and the 1.40625-degree model, as shown in \cref{fig:degree}. For the 5.625-degree model, the input global data is represented as a matrix of size [32, 64], while the input matrix for the 1.40625-degree model is of size [128, 256]. A smaller grid implies higher data resolution and more information. The 1.4 version model achieves nearly a 10\% improvement at 2-3 week intervals and a smaller 4-6\% improvement for 4-6 week periods.

\section{Discussion and Conclusion}
Our work significantly advances the field of subseasonal-to-seasonal (S2S) weather forecasting by demonstrating that a multi-stage optimization approach can substantially enhance the prediction skill of deep learning methods, which were previously considered less effective than climatology\cite{nathaniel2024chaosbench}. The proposed method improves key skill metrics, outperforming SOTA ECMWF-S2S by 19-91\%. The study challenges traditional forecasting assumptions by showing that our framework can markedly improve rolling forecast performance. It offers valuable insights and theoretical analyses, elucidating the core issues in rolling forecasting and how our teacher-forcing design addresses them. Our model-agnostic framework could become the general training approach for various model designs, and we anticipate that incorporating more features and enhanced neural network architectures will rapidly improve S2S forecasting results in the near future.

{\small
\bibliographystyle{ieee_fullname}
\bibliography{egbib}
}
\newpage
\clearpage
\appendix
\section*{Appendix} 
\addcontentsline{toc}{section}{Appendix} 
\section{Resource Open Sourcing}
Our method and codes are open-sourced at \url{https://anonymous.4open.science/r/Baguan-S2S-23E7/}. As part of the Baguan weather and climate model series, we will have more forthcoming work and results to be released. We also plan to make the model weights publicly available.

\section{Proof of Our Statement}
\label{appendix:proof}

\newtheorem{thm}{Theorem}
\newtheorem{prop}{Proposition}
\newtheorem{lemma}{Lemma}
\newtheorem{cor}[thm]{Corollary}
\newtheorem{definition}[thm]{Definition}

\def \y {\mathbf{y}}
\def \E {\mathrm{E}}
\def \x {\mathbf{x}}
\def \g {\mathbf{g}}
\def \L {\mathcal{L}}
\def \D {\mathcal{D}}
\def \z {\mathbf{z}}
\def \u {\mathbf{u}}
\def \H {\mathcal{H}}
\def \w {\mathbf{w}}
\def \R {\mathbb{R}}
\def \S {\mathcal{S}}
\def \regret {\mbox{regret}}
\def \Uh {\widehat{U}}
\def \Q {\mathcal{Q}}
\def \W {\mathcal{W}}
\def \N {\mathcal{N}}
\def \A {\mathcal{A}}
\def \q {\mathbf{q}}
\def \v {\mathbf{v}}
\def \M {\mathcal{M}}
\def \c {\mathbf{c}}
\def \ph {\widehat{p}}
\def \d {\mathbf{d}}
\def \p {\mathbf{p}}
\def \q {\mathbf{q}}
\def \db {\bar{\d}}
\def \dbb {\bar{d}}
\def \I {\mathcal{I}}
\def \f {\mathbf{f}}
\def \a {\mathbf{a}}
\def \b {\mathbf{b}}
\def \ft {\widetilde{\f}}
\def \bt {\widetilde{\b}}
\def \h {\mathbf{h}}
\def \B {\mathbf{B}}
\def \bts {\widetilde{b}}
\def \fts {\widetilde{f}}
\def \Gh {\widehat{G}}
\def \bh {\widehat{b}}
\def \fh {\widehat{f}}
\def \vb {\bar{v}}
\def \zt {\widetilde{\z}}
\def \zts {\widetilde{z}}
\def \s {\mathbf{s}}
\def \gh {\widehat{\g}}
\def \vh {\widehat{\v}}
\def \Sh {\widehat{S}}
\def \rhoh {\widehat{\rho}}
\def \hh {\widehat{\h}}
\def \C {\mathcal{C}}
\def \V {\mathcal{V}}
\def \t {\mathbf{t}}
\def \xh {\widehat{x}}
\def \Ut {\widetilde{U}}
\def \wt {\widetilde{\w}}
\def \Th {\widehat{T}}
\def \Ot {\tilde{\mathcal{O}}}
\def \X {\mathcal{X}}
\def \nb {\widehat{\nabla}}
\def \K {\mathcal{K}}
\def \P {\mathbb{P}}
\def \T {\mathcal{T}}
\def \F {\mathcal{F}}
\def \ft{\widetilde{f}}
\def \xt {\widetilde{x}}
\def \Rt {\mathcal{R}}
\def \V {\mathcal{V}}
\def \Rb {\bar{\Rt}}
\def \wb {\bar{\w}}
\def \fh {\widehat{f}}
\def \wh {\widehat{\w}}
\def \lh {\widehat{\lambda}}
\def \e {\mathbf{e}}
\def \B {\mathcal{B}}
\def \P {\mathcal{P}}
\def \vb {\bar{v}}
\def \ub {\bar{u}}
\def \Rt {\mathcal{R}}
\def \wh {\widehat{w}}
\def \Lh {\widehat{\L}}
\def \rh {\widehat{r}}
\def \G {\mathcal{G}}
\def \qh {\widehat{q}}
\def \Qh {\widehat{Q}}
\def \Gh {\widehat{G}}
\def \Ph {\widehat{P}}
\def \U {\mathcal{U}}
\def \diag {\mbox{diag}}
\def \qt {\widetilde{q}}
\def \sh {\widehat{s}}
\def \st {\widetilde{s}}
\def \ut {\widetilde{u}}
\def \yt {\widetilde{y}}
\def \vb {\bar{v}}
\def \ub {\bar{u}}
\def \Rt {\mathcal{R}}
Deep linear model refers to a prediction $g(x) = \prod_{i=1}^L \Theta_i x$, where $x \in \R^d$ and $\Theta_i \in \R^{d\times d}, \forall i \in [L]$. The goal is to find $\Theta_i, i \in [L]$ such that $\ell(\Theta_{1:L}) = \E_{(x,y)\sim\P}\left[|y - g(x)|^2\right]$ is minimized, where $y \in \R^d$~\footnote{$y$ can be vector of any dimension. Here we set the dimension of $y$ to be $d$ just for the convenience of discussion because it results in square matrices for all $\Theta_i$}. When $y = \Phi x$ and $x$ is sampled from a normal distribution $\N(0, I)$. The objective function is simplified as $\ell(\Theta_{1:L}) = |\Phi - \Theta_{1:L}|^2_2$. In this note, we consider a special case of deep linear model that has $\Theta_i = \Theta, i \in [L]$. It fits in better with the rolling out method in our study where the same model is used at every step of rolling out. We will show that in order to achieve a fast convergence of deep linear model with all $\Theta_i$ being the same, the initial solution $\Theta^{(0)}$ needs to have a small loss $\ell(\Theta^{(0)})$. In particular, the deeper the linear model is, the smaller the initial loss $\ell(\Theta^{(0)})$ is. It helps explain why directly training a rolling out network with a long horizon can be unstable, and why a curriculum learning type approach is preferred. Our analysis follows closely the work ``Gradient descent with identity initialization efficiently learns positive definite linear transformations by deep residual networks''. 

First, it is easy to verify that Lemma 3 from the original paper remains unchanged while Lemma 2 should be modified as follows due to our special setup that $\Theta_i = \Theta, i \in [L]$
\begin{eqnarray}
|\nabla \ell(\Theta)|^2 \geq 4L^2\ell(\Theta)\left(1 - \sigma_{\min}(\Theta)\right). \label{eqn:loss}
\end{eqnarray}
This change will lead to the following change to inequality in (4) from the original work, i.e.,
\[
\ell(t+1) \leq \left(1 - 2\eta L^2(1 - \Rt(t))^L \right) \ell(t)
\]
under the condition
\begin{eqnarray}
\eta \leq \frac{1}{3Ld^5\max\left\{(1 + \R(t+1))2L, |\Phi|_2 \right\}}, \label{eqn:eta-1}
\end{eqnarray}
where $\ell(t) = \ell(\Theta^{(t)})$ and $\Rt(t) = |\Theta - I|_2$. In addition, due to our special setup, the inequality (2) from the original paper will be modified as
\begin{eqnarray}
\Rt(t+1) \leq \Rt(t) + \eta L(1 + \Rt(t))^L \sqrt{\ell(t)}. \label{eqn:rt}
\end{eqnarray}
Inequalities in (\ref{eqn:loss}) and (\ref{eqn:rt}) form the basis for the overall analysis. 

Second, given the changed inequalities in (\ref{eqn:loss}) and (\ref{eqn:rt}), it is not difficult to see that by simply redefining $\eta' = \eta L$, we can follow the identical analysis from the original work. As a result, we will need to change the condition for $\eta$ from (\ref{eqn:eta-1}) to the following form
\begin{eqnarray}
\eta \leq \frac{1}{3L^2d^5\max\left\{c^4, |\Phi|_2 \right\}}. \label{eqn:eta-2}
\end{eqnarray}
Following the same analysis as the original work, under the condition (\ref{eqn:eta-2}) and 
\begin{eqnarray}
\ell(0) \leq \frac{\ln c}{4c^{10}}, \label{eqn:init}
\end{eqnarray}
where $c \leq \exp(3L/4)$, we have
\begin{eqnarray}
\ell(t) \leq \exp\left(-\frac{2t\eta L}{c^4}\right) \ell(0). \label{eqn:iter-2}
\end{eqnarray}
By assuming that the target matrix $\Phi$ is sufficiently large, i.e.,
\[
|\Phi|_2 \geq c^4,
\]
we have
\begin{eqnarray}
\ell(t) \leq \exp\left(-\frac{2t}{3Ld^5c^9}\right) \ell(0). \label{eqn:iter-2}
\end{eqnarray}
As indicated by (\ref{eqn:iter-2}), to achieve a faster convergence for deep linear model, we need $c$ to be sufficiently small, while a sufficiently small $c$ will result in a small initial error $\ell(0)$. This is because the upper bound in (\ref{eqn:init}), i.e., $\ln c/c^{10}$, is a monotonically increasing function when $c \in [1, e^{0.1}]$. Hence, a smaller $c$ in the range between $1$ and $e^{0.1}$ will result a smaller initial error $\ell(0)$. Hence, to obtain a reliable training for a deep linear model, we need to choose an initial solution with a small error. 

\section{Implementation of 6 Sets of Adapters}
\label{appendix:adapter}
\begin{figure*}[th]
    \centering
    \includegraphics[width=0.8\linewidth]{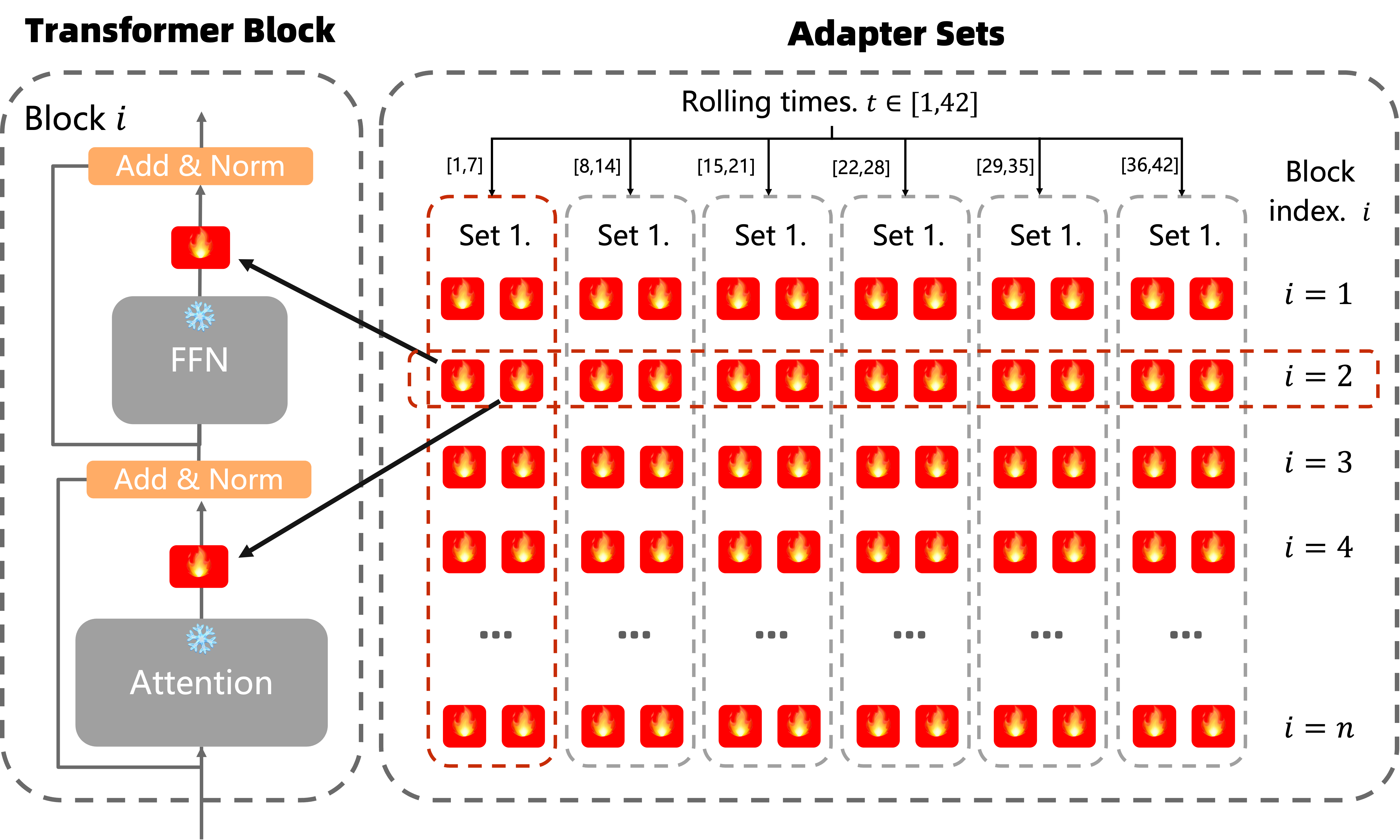} 
    \caption{Implementation of 6 sets of adapter.}
    \label{fig:adx-adp}
\end{figure*}

We employed an adapter-based PEFT method as illustrated in \cref{fig:adx-adp}. For each transformer block in the model, we added an adapter for fine-tuning while keeping all the original parameters of the model frozen. Each adapter is a small MLP network that consists of two fully connected layers with an activation layer in between. The hidden dimension used for the adapters is consistent with the hidden dimension of the transformer blocks, both set to 1024. In the 42 iterations of rolling prediction, we utilized six sets of adapters. Specifically, during the first seven iterations, we used adapter set 1; from the eighth to the fourteenth iteration, we employed adapter set 2, and so on. During training, all adapters were trainable. The parameter count of a single adapter is low and the parameter count of each adapter set constitutes 4\% of the total model parameters, allowing for constraints on the model training.

\section{Ablation of Different Backbones}
\label{appendix:backbone}
\begin{figure}[th]
    \centering
    \includegraphics[width=1\linewidth]{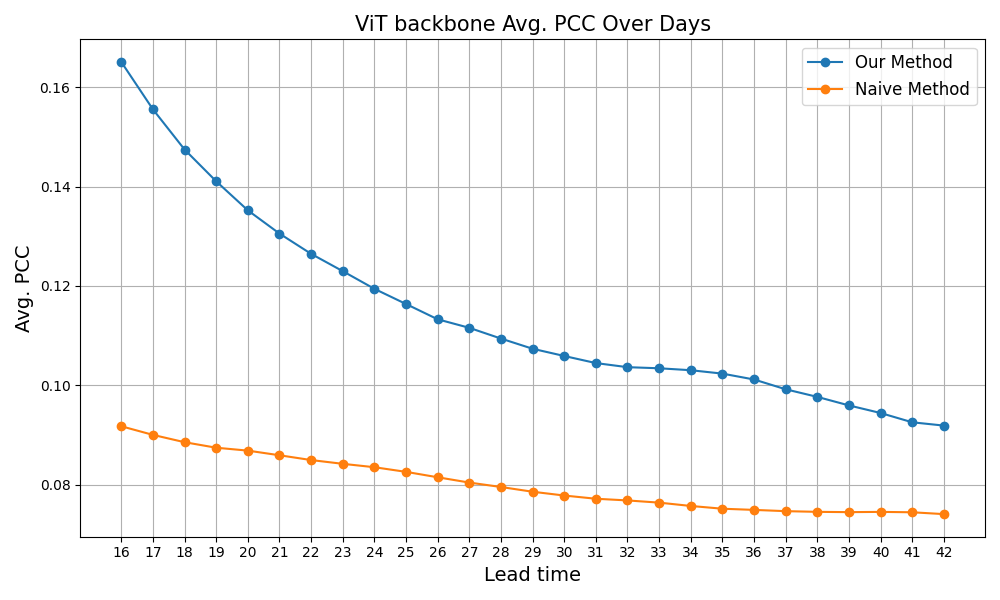} 
    \caption{PCC comparison between our method and naive method.}
    \label{adx:vit}
\end{figure}

\begin{figure}[th]
    \centering
    \includegraphics[width=1\linewidth]{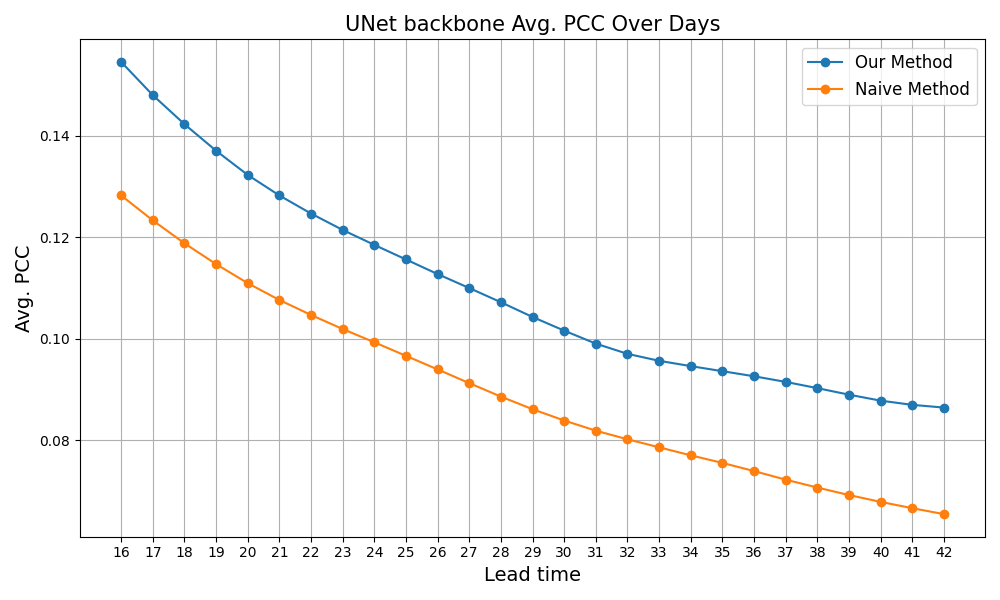} 
    \caption{PCC comparison between our method and naive method.}
    \label{adx:unet}
\end{figure}

We compared the performance of our method and the naive method using ViT and UNet backbones, as shown in \cref{adx:vit} and \cref{adx:unet}.
\section{5.625-degree Results}
\label{appendix:5.625}
\begin{figure*}[th]
    \centering
    \includegraphics[width=1\linewidth]{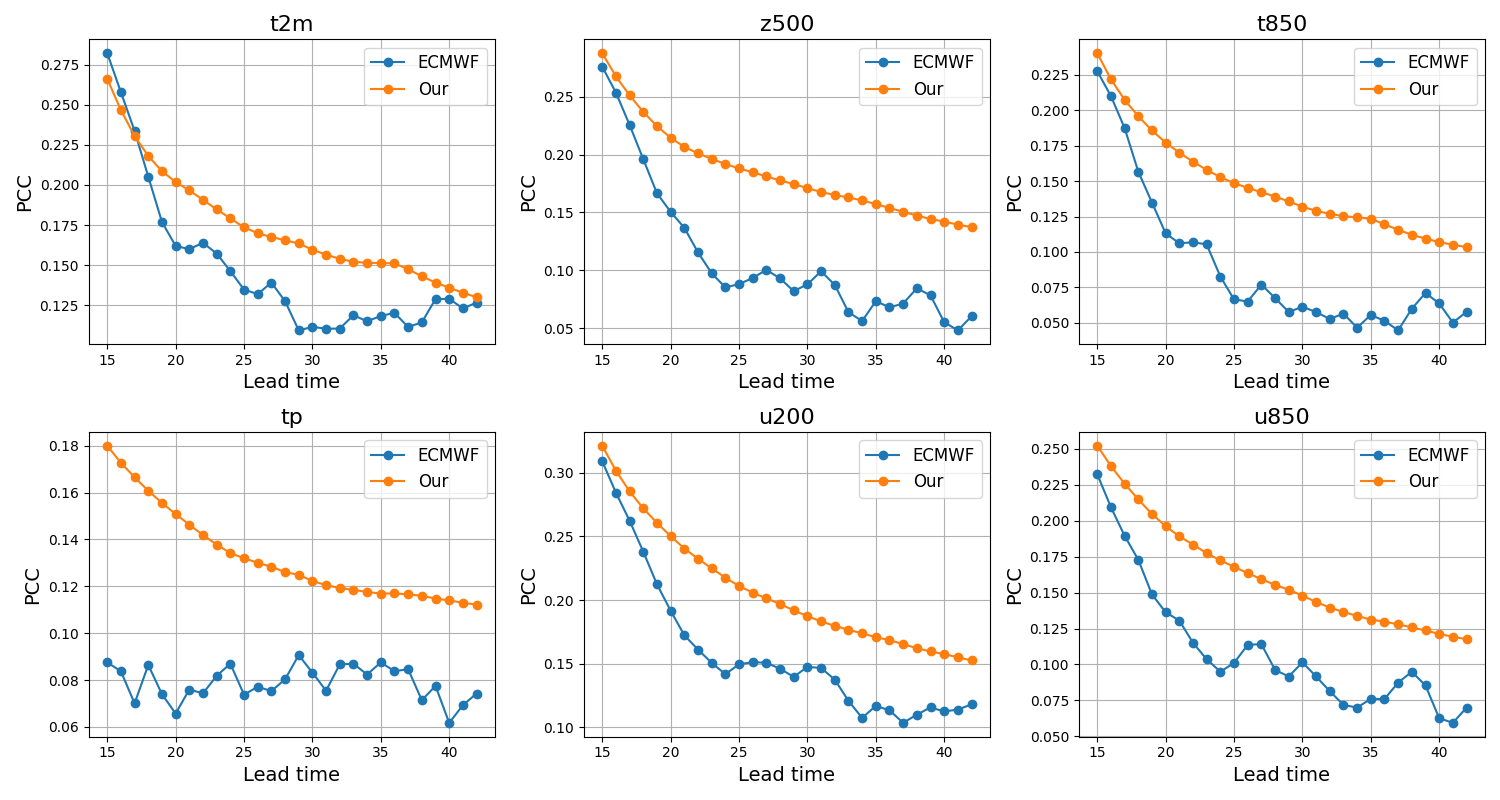} 
    \caption{PCC comparsion between ECMWF-S2S and Ours (5.625 degree).}
    \label{fig:adx-pcc}
\end{figure*}

\begin{figure*}[th]
    \centering
    \includegraphics[width=1\linewidth]{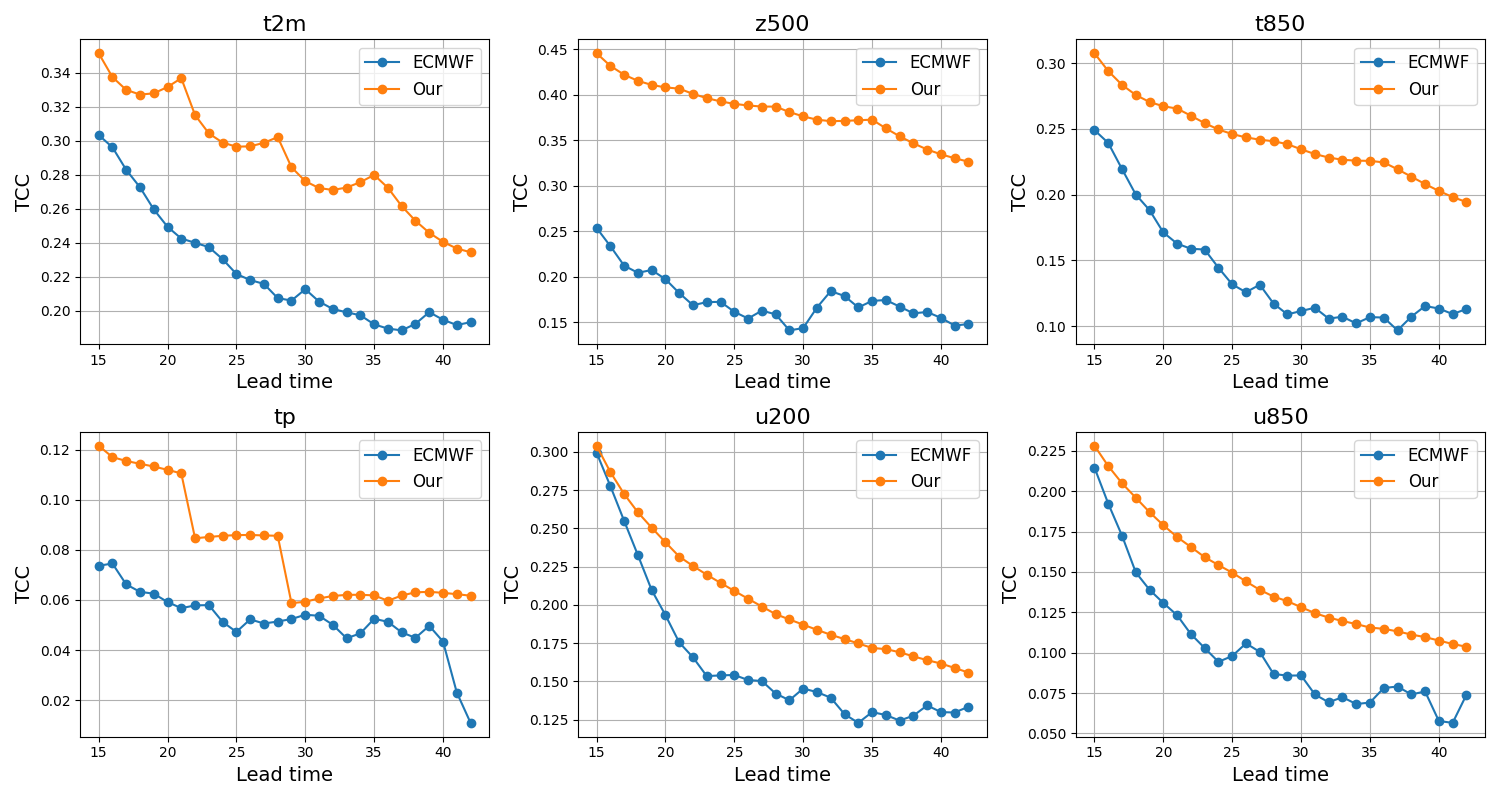} 
    \caption{TCC comparsion between ECMWF-S2S and Ours (5.625 degree).}
    \label{fig:adx-tcc}
\end{figure*}

The model prediction results at 5.625 degrees compared with ECMWF are shown in \cref{fig:adx-pcc} and \cref{fig:adx-tcc}. The training process for the 5.625-degree model is identical to that of the 1.40625-degree model described in the main text. The only difference is that the training data has been switched to the 5.625-degree resolution. Our model can surpass the ECMWF-S2S with a resolution of 1.5 degree, even at a resolution of 5.625 degree.

\section{Visualization}
\label{appendix:Visualization}
In the S2S task, tp, z500, and u200 are of significant value for predicting storms, rainfall, and temperature. We conducted global data visualization for these parameters. From the figure, it is evident that our prediction results have a lower error compared to ECMWF.
\begin{figure}[th]
    \centering
    \includegraphics[width=1\linewidth]{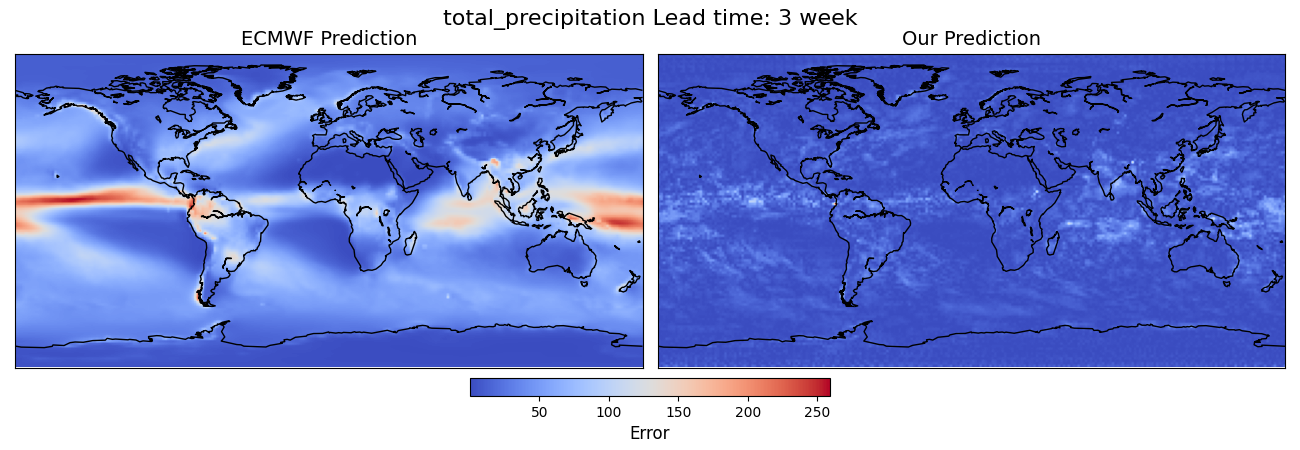} 
    \includegraphics[width=1\linewidth]{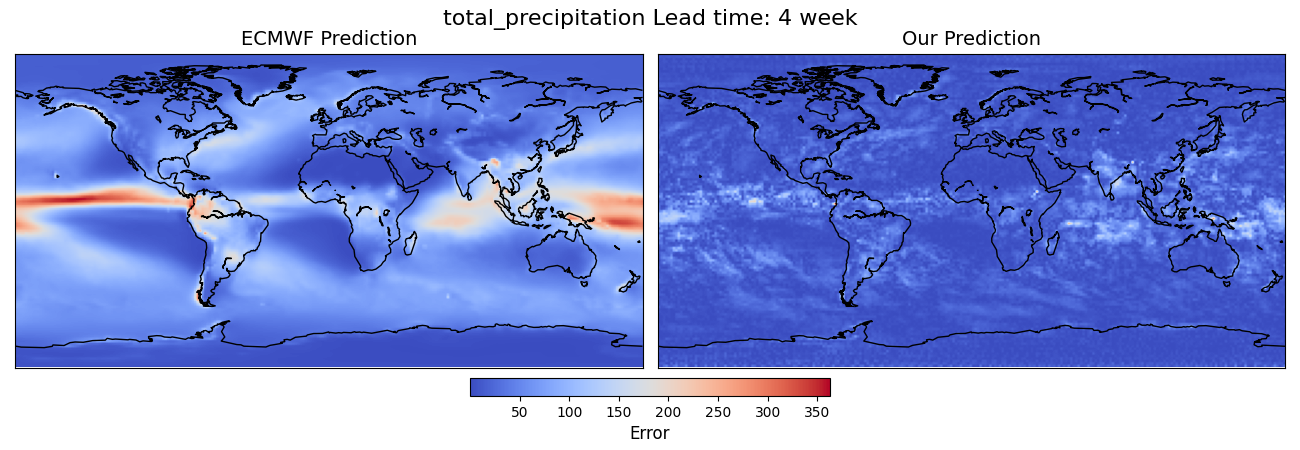} 
    \includegraphics[width=1\linewidth]{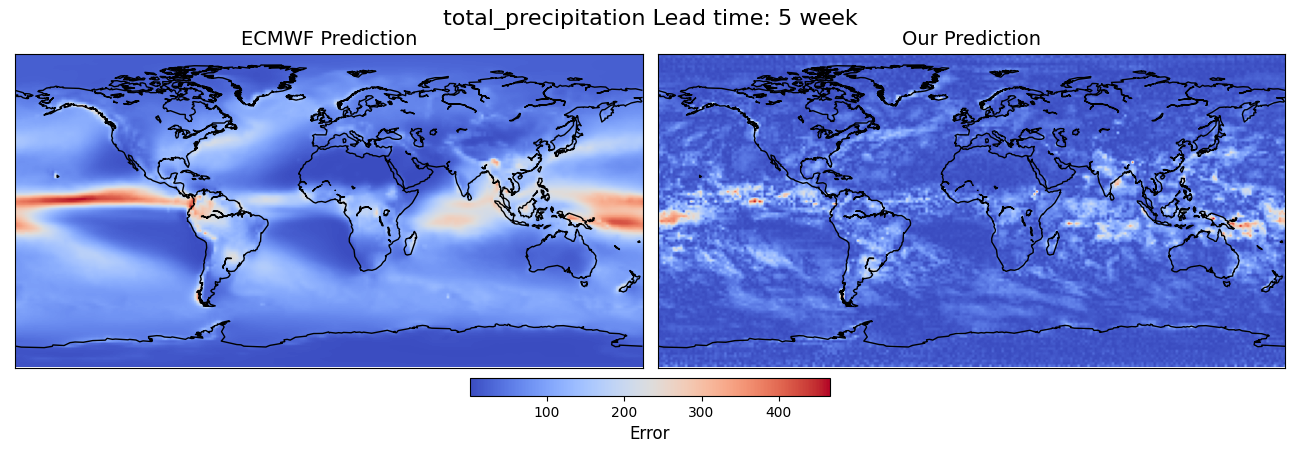} 
    \includegraphics[width=1\linewidth]{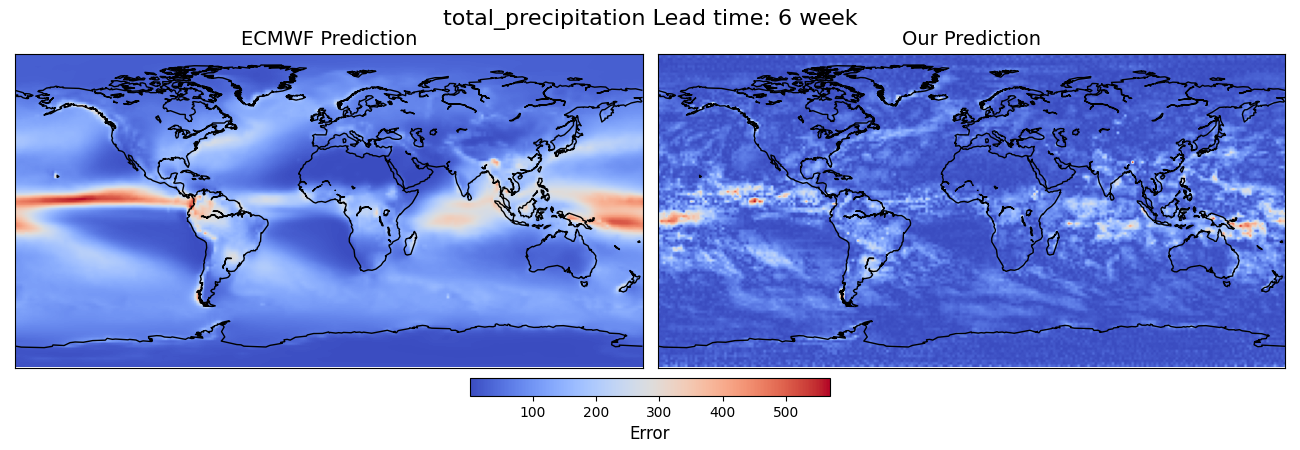} 
    \caption{tp visualized error comparison between our method and ECMWF.}
    \label{adx:vis-z500}
\end{figure}

\begin{figure}[th]
    \centering
    \includegraphics[width=1\linewidth]{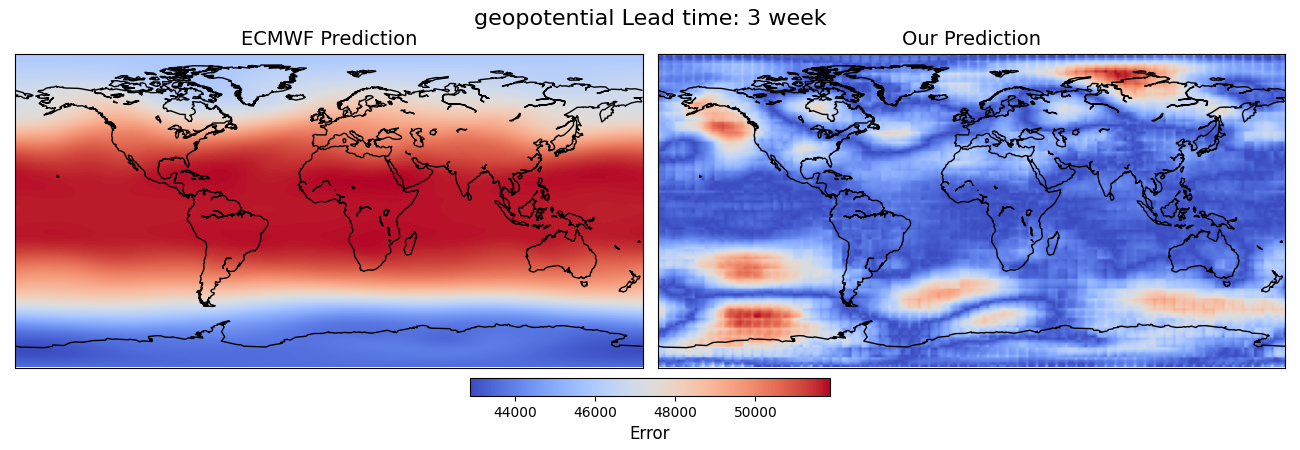} 
    \includegraphics[width=1\linewidth]{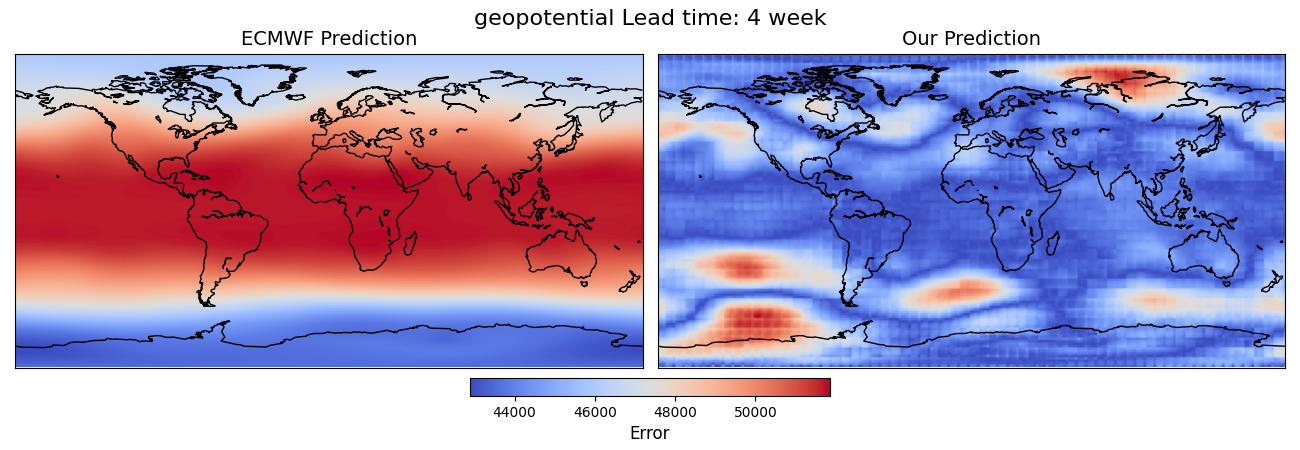} 
    \includegraphics[width=1\linewidth]{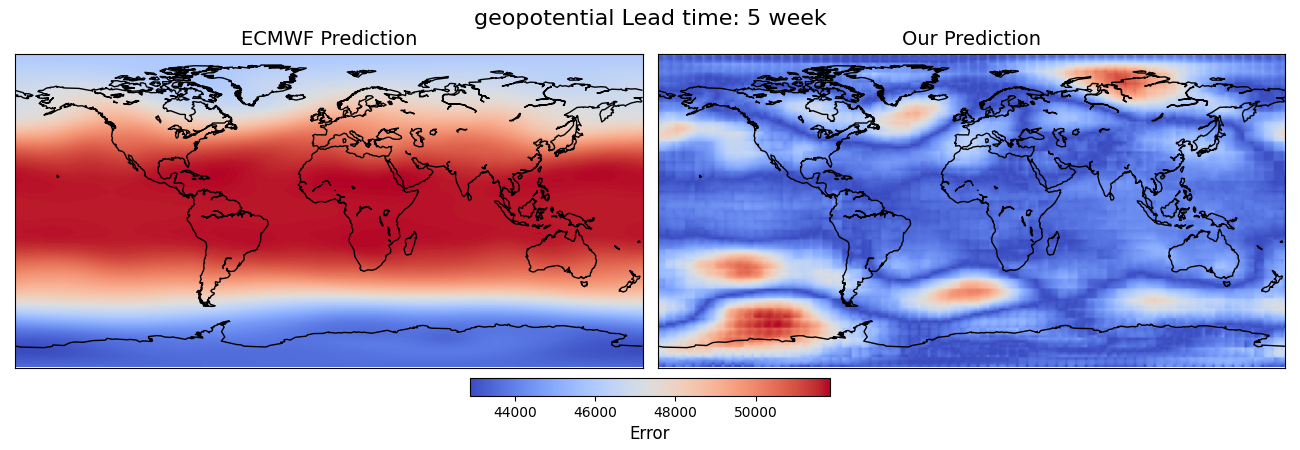} 
    \includegraphics[width=1\linewidth]{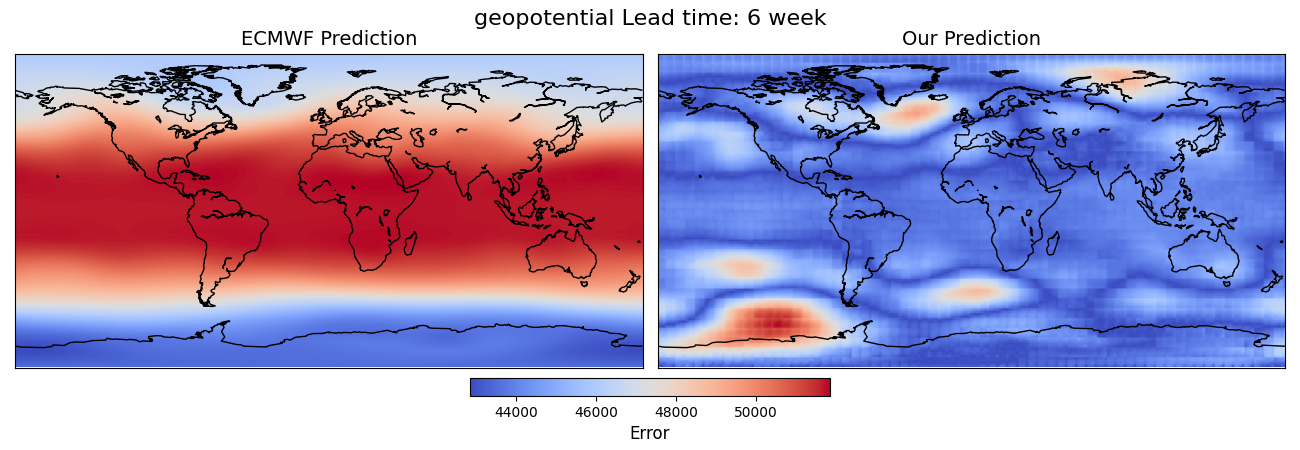} 
    \caption{z500 visualized error comparison between our method and ECMWF.}
    \label{adx:vis-z500}
\end{figure}

\begin{figure}[th]
    \centering
    \includegraphics[width=1\linewidth]{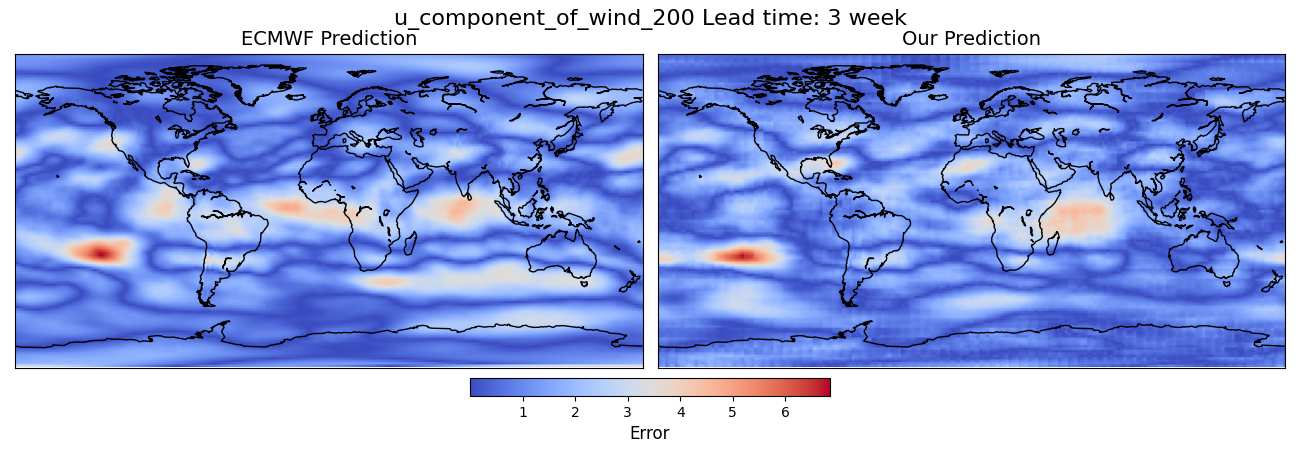} 
    \includegraphics[width=1\linewidth]{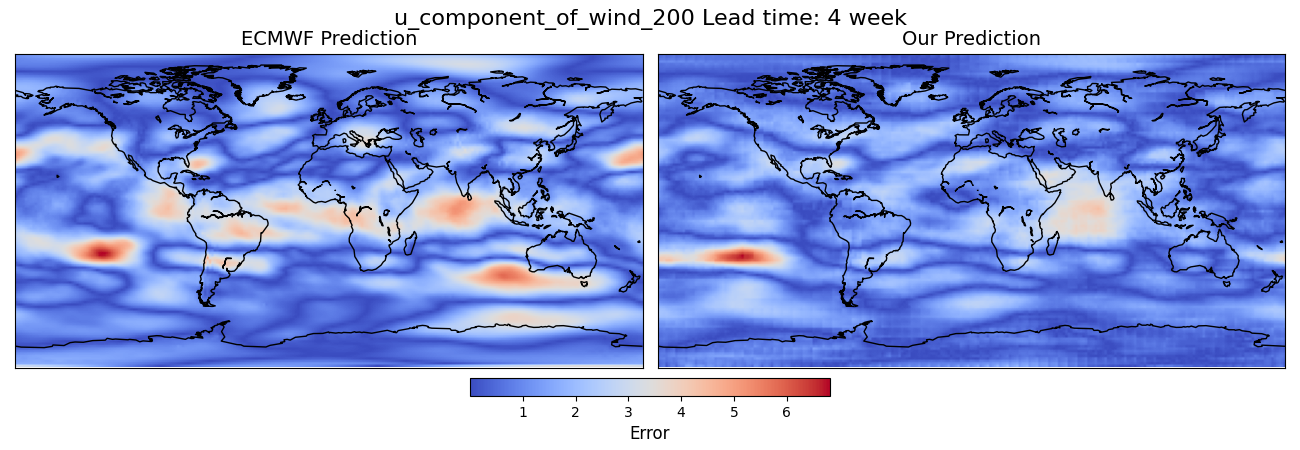} 
    \includegraphics[width=1\linewidth]{figures/Leadtime4_u_component_of_wind_200.png} 
    \includegraphics[width=1\linewidth]{figures/Leadtime4_u_component_of_wind_200.png} 
    \caption{u200 visualized error comparison between our method and ECMWF.}
    \label{adx:vis-u200}
\end{figure}

\section{Detailed Results of the Ablation Study}
In the ablation experiment section of the main text, we present the curves and results of the average PCC. Here, we will provide the detailed PCC results for each individual ablation experiment. It is important to note that the models used in the ablation experiments are all at a resolution of 5.625 degrees, which presents certain disadvantages compared to the 1.40625-degree models.

\begin{figure*}[th]
    \centering
    \includegraphics[width=1\linewidth]{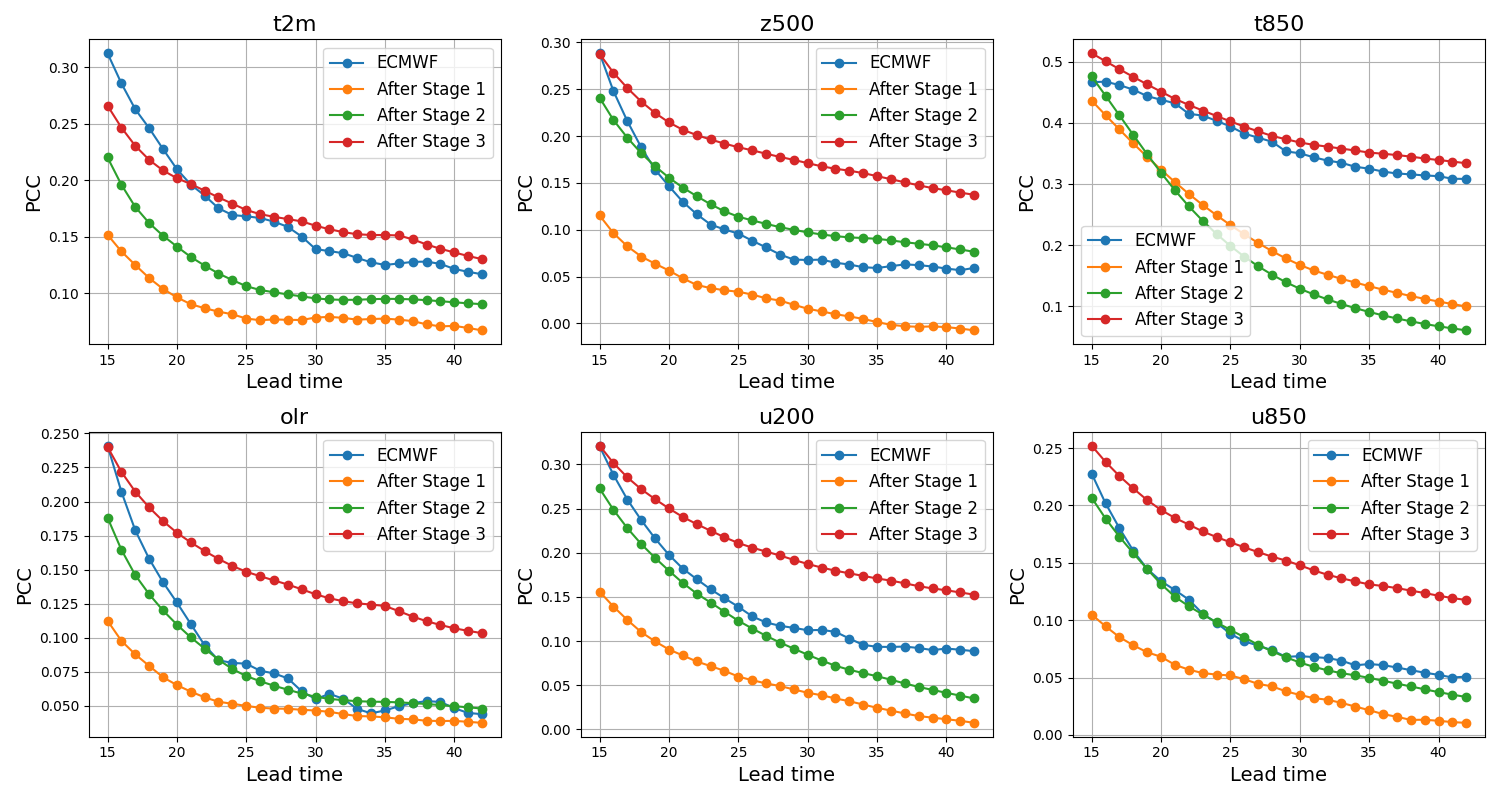} 
    \caption{PCC comparsions between ECMWF-S2S and Ours in three stages.}
\end{figure*}

\begin{figure*}[th]
    \centering
    \includegraphics[width=1\linewidth]{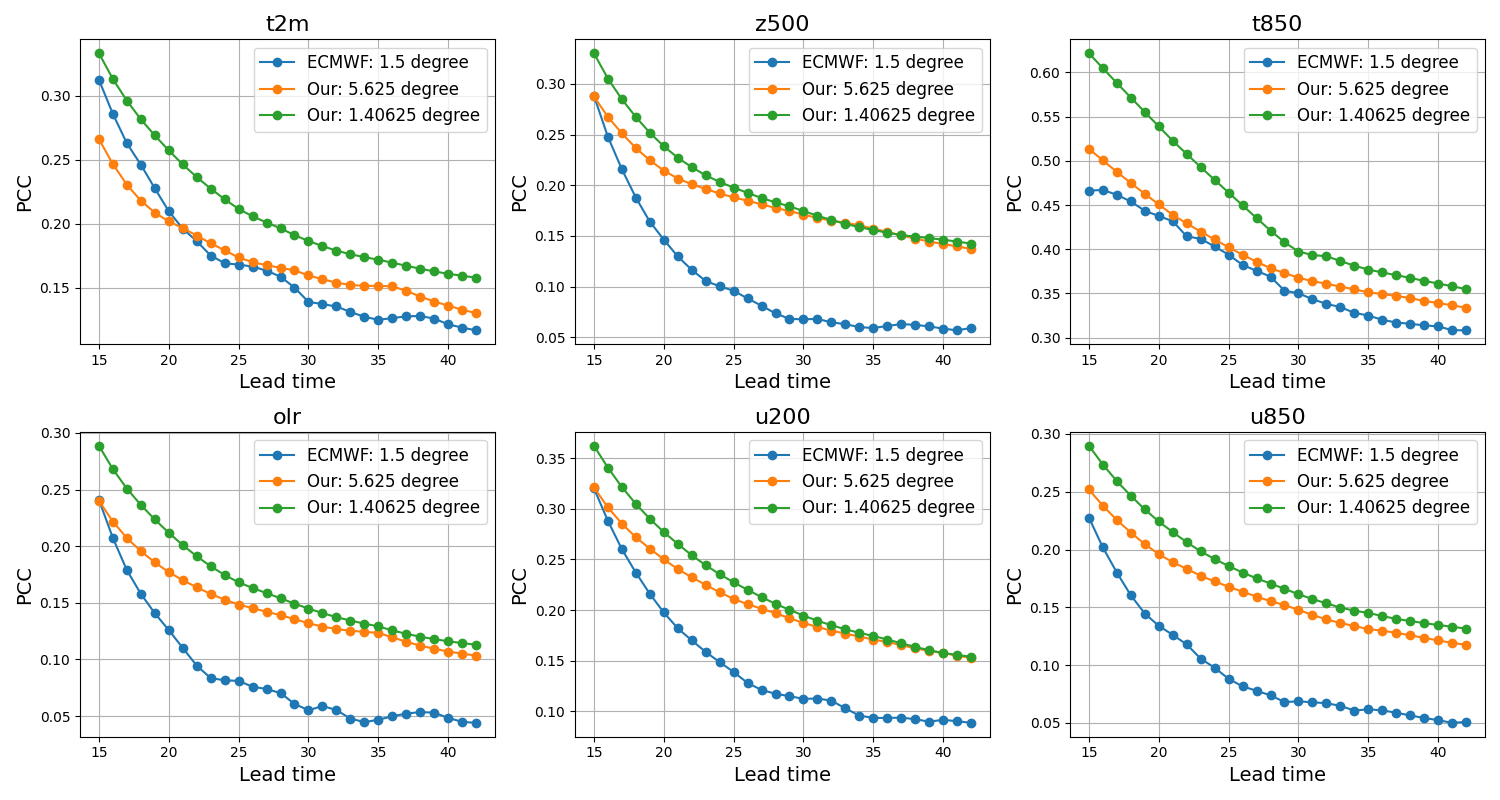} 
    \caption{PCC comparsions between Our model trained with 5.625-degree and 1.40625-degree.}
\end{figure*}

\begin{figure*}[th]
    \centering
    \includegraphics[width=1\linewidth]{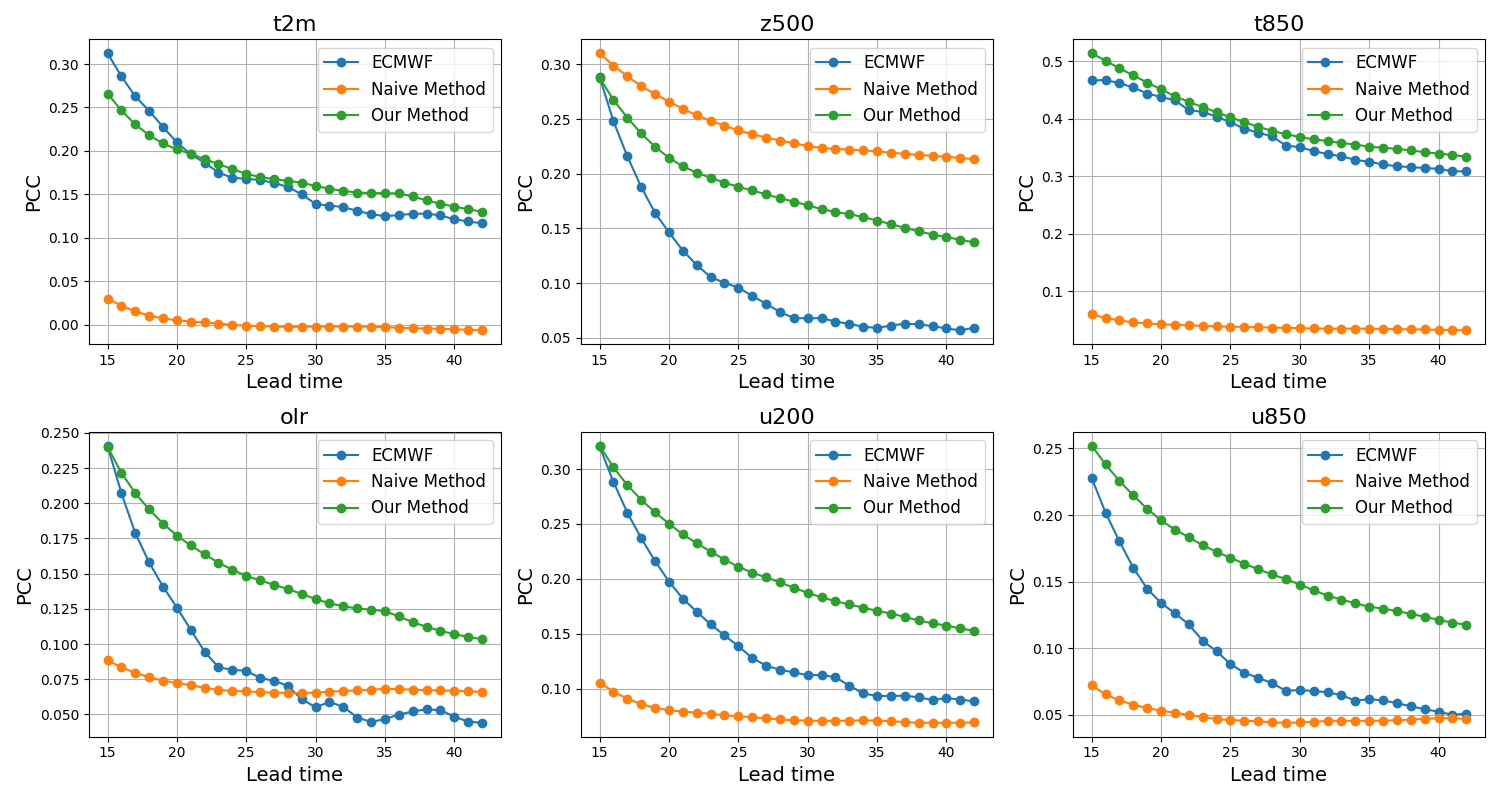} 
    \caption{PCC comparsions between Our method and Naive method.}
\end{figure*}

\begin{figure*}[th]
    \centering
    \includegraphics[width=1\linewidth]{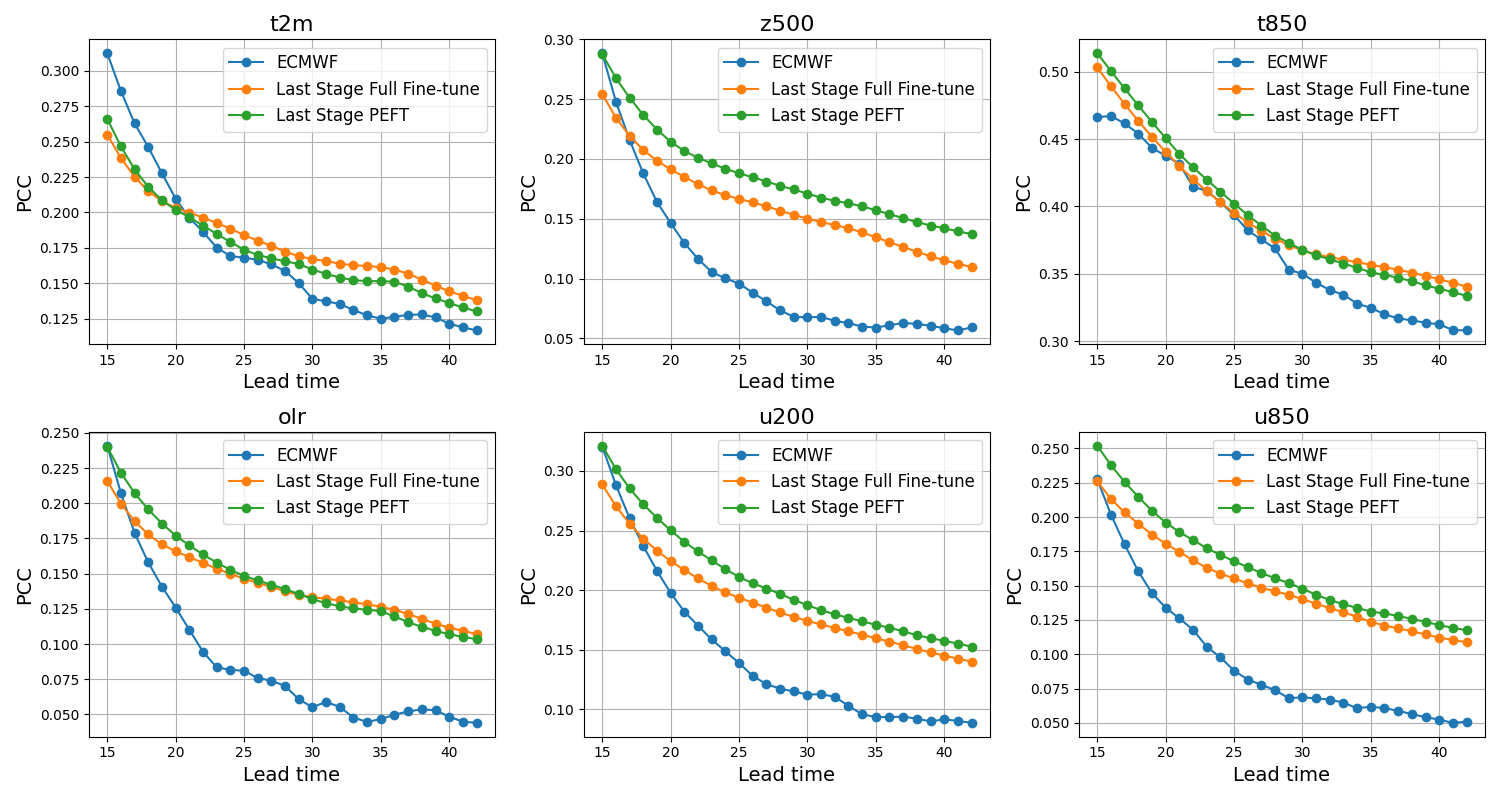} 
    \caption{PCC comparsions between last stage full fine-tuning and PEFT.}
\end{figure*}

\end{document}